\documentclass{article} %
\usepackage{iclr2027_conference,times}

\usepackage{amsmath,amsfonts,bm}

\def\eqref#1{equation~\ref{#1}}

\def\1{\bm{1}}

\DeclareMathAlphabet{\mathsfit}{\encodingdefault}{\sfdefault}{m}{sl}
\SetMathAlphabet{\mathsfit}{bold}{\encodingdefault}{\sfdefault}{bx}{n}

\usepackage{hyperref}
\usepackage{url}
\usepackage{amsmath,amssymb,booktabs,array,tabularx,multirow,graphicx}
\usepackage{enumitem}
\usepackage{colortbl}
\newcommand{\name}{\textsc{Uruqi}}

\newenvironment{packeditemize}{
\begin{list}{$\bullet$}{
\setlength{\labelwidth}{8pt}
\setlength{\itemsep}{0pt}
\setlength{\leftmargin}{\labelwidth}
\addtolength{\leftmargin}{\labelsep}
\setlength{\parindent}{0pt}
\setlength{\listparindent}{\parindent}
\setlength{\parsep}{0pt}
\setlength{\topsep}{3pt}}}{\end{list}}
\newcommand{\na}{--}

\title{Uruqi: Learning Spatial Cognition from \\Visual Experience}

\author{
Shichao Li\textsuperscript{1}
\quad
Meiqi Wang\textsuperscript{2}
\quad
Fei Su\textsuperscript{1}
\quad
Zhicheng Zhao\textsuperscript{1}
\\[0.4em]
\normalfont
\textsuperscript{1}Beijing University of Posts and Telecommunications
\qquad
\textsuperscript{2}Tsinghua University
\\[0.25em]
\texttt{lishichao@bupt.edu.cn}
\quad
\texttt{zhaozc@bupt.edu.cn}
}

\iclrfinalcopy %
\begin{document}

\maketitle

\begin{abstract}
Spatial intelligence requires maintaining a coherent understanding of the world as the embodied agent moves. 
Like humans, the agent must use its own motion to interpret changes across observations and update object locations and spatial relations accordingly.
Despite spatial post-training having substantially broadened the spatial intelligence of vision-language models (VLMs), they still struggle with two atomic spatial capabilities: tracking self-motion and mapping the surrounding world during motion.
To address this gap, we provide dense multi-turn supervision over interleaved atomic capabilities within each training episode, mimicking the visual experience of a continuously moving agent that reasons as it observes.
To scale this up, we synthesize 11,738 motif-driven camera trajectories over a broad range of 3D scenes, supporting self-motion tracking, persistent object mapping, and rich spatial operations within each visual experience.
By training models to reason over these atomic questions, our \textsc{URUQI}$_{\mathrm{Syn}}$-8B improves accuracy from 15.84\% to 47.73\% on our ~\name{} benchmark comprising 52k questions across 2.7k episodes. \textsc{URUQI}-SI-Mix-8B further reaches 50.41\%,
comparable to the 50.08\% achieved by GPT-6 Astra.
Trained solely on our synthesized data, \textsc{URUQI}$_{\mathrm{Syn}}$-8B achieves an average
relative accuracy improvement of 17.13\% over its InternVL3-8B backbone across three external spatial benchmarks. These results highlight continuous visual experience as a
scalable source of supervision for developing spatial
cognition in VLMs.

\end{abstract}

\section{Introduction}

Spatial reasoning is becoming a core capability of general-purpose vision-language models (VLMs) that aim to understand the physical world. 
A VLM with native spatial intelligence is expected to ground open-vocabulary objects and integrate metric, relational, and reference-frame spatial information directly into open-ended visual-language reasoning.
Recent work has shown that such capabilities can be substantially improved with large-scale spatial instruction tuning \citep{chen2024spatialvlm, cheng2024spatialrgpt, cai2026scaling}.

\begin{figure*}[t]
    \centering
    \includegraphics[width=\linewidth]{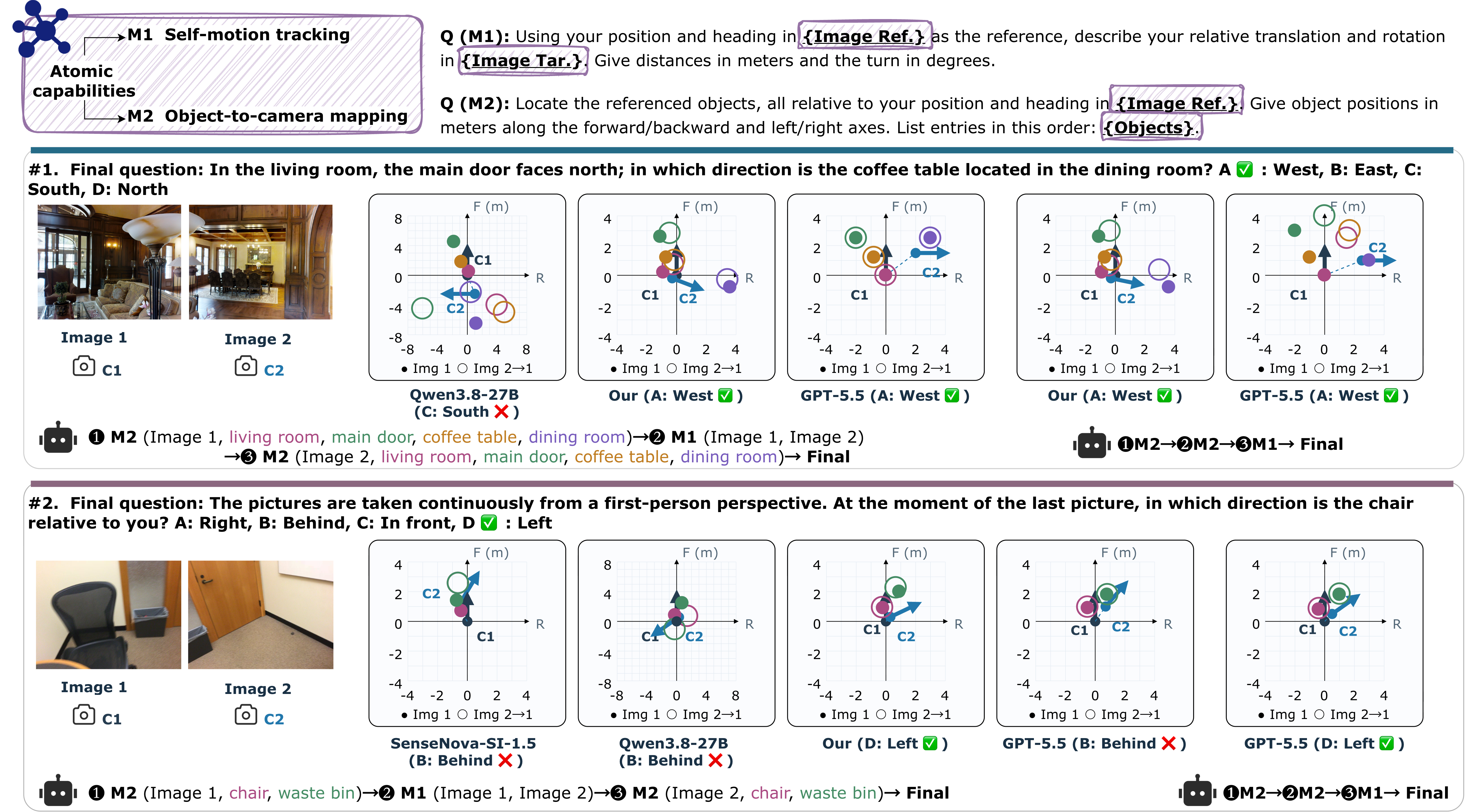}

    \caption{Visualization of two MMSI-Bench~\citep{yang2026mmsi} examples by querying M1 and M2 before asking the final question. The query order is shown below each plot. The coordinate frame is defined at $C_1$. The black and blue arrows mark the viewing directions at $C_1$ and $C_2$, respectively, while the blue dashed line denotes the displacement from $C_1$ to $C_2$ predicted by M1. For M2, filled markers show object locations predicted from Image~1, while hollow markers show Image~2 predictions transformed into the $C_1$ frame using the predicted self-motion. The overlap between filled and hollow markers reflects the cross-view consistency of object-location predictions.}
    \label{fig:overview}
    \vspace{-2ex}
\end{figure*}

However, it remains unclear whether success on diverse spatial questions reflects a coherent understanding of the scene across changing viewpoints. Figure~\ref{fig:overview} illustrates this issue with two examples from MMSI-Bench~\citep{yang2026mmsi} through self-motion tracking (M1) and object mapping (M2). Qwen3.8-27B~\citep{qwen2026qwen38_27b} and SenseNova-SI-1.5~\citep{cai2026scaling} make inaccurate motion and object-location predictions, while their estimates of the same objects are inconsistent across viewpoints. GPT-5.5~\citep{openai2026gpt55} also predicts inaccurate object locations, but its estimates remain consistent across views. We swap the query order of M1 and M2 and find that its object predictions change accordingly. This dependency on query order suggests that GPT-5.5 computes object locations from earlier location estimates and its predicted camera motion. These results highlight a mismatch between task accuracy and coherent spatial understanding across observations.

To address this gap, prior work has explored supervising intermediate reasoning steps through chain-of-thought~\citep{spatialcot,cai2026scaling,li2026spatialladder}, cognitive maps~\citep{wang2025mindcube,yang2025thinking}, structured representations~\citep{hua2026unleashing}, and executable reasoning programs~\citep{marsili2025visual}. 
However, linguistic reasoning traces provide potentially ambiguous supervision for geometric state transitions~\citep{kancheti2026chain}, while explicit maps and programs rely on predefined representations and task-specific operations. 
This leaves a more general question: \emph{What intermediate supervision can capture how spatial information evolves across observations and generalize to unseen spatial tasks?}

We answer this question from a basic geometric property of \emph{visual experience}: consecutive spatial observations are connected by the self-motion of the observer~\citep{wang2000updating,wolbers2008spatial}. Unlike generic multi-image understanding, which mainly integrates semantic information across images~\citep{jiang2024mantis}, spatial reasoning must also recover the geometric relationship between observations. Our key insight is that observer self-motion is more than a prediction target because it determines how spatial information should change from one observation to the next.

This suggests that motion estimation should be explicitly coupled with spatial state updates during learning. The model must first recover how the observer moves, use self-motion to update spatial relations established from previous observations, and then use the resulting state to answer the required query. Current spatial post-training often supervises these capabilities through separate task-specific questions, leaving the transition from observer motion to spatial state update largely implicit.

Learning this transition requires visual observations to be aligned with observer motion and the resulting changes in spatial relations.
We therefore introduce \name{}, a scalable spatial compiler for synthesizing continuous visual experiences and compiling them into dense spatial supervision.
\name{} generates motif-driven camera trajectories across diverse 3D scenes, where each motif controls how observer motion, visibility, and spatial evidence unfold over time. Using the underlying camera poses, scene geometry, and object identities, each trajectory is then compiled into multi-turn supervision over three linked capabilities: \textbf{self-motion tracking}, which captures changes in viewpoint; \textbf{persistent object mapping}, which tracks object locations as the observer moves; and \textbf{operations over state}, which supports downstream reasoning over the acquired spatial information. Questions derived from the same trajectory are further organized into shared episodes, so that motion, state updates, and spatial operations are learned as connected parts of the same experience rather than as isolated QA examples.

To the best of our knowledge, \name{} is the first to jointly supervise self-motion, persistent object mapping, and downstream spatial reasoning within continuous visual experiences. Scaling this compilation across diverse trajectories, we construct ~\name{}-600k, a large-scale dataset with dense spatial supervision. {\textsc{URUQI}$_{\mathrm{Syn}}$-8B} trained on this corpus achieves substantial gains in episodic spatial reasoning on \name{} benchmark in Section ~\ref{sec:indomain} and transfers effectively to external spatial benchmark without using any training data from these target benchmarks (Section ~\ref{sec:exp_transfer}). Moreover, combining our supervision with only 100K Sensenova-SI-8M examples further improves SenseNova-SI-1.5~\citep{cai2026scaling}, showing that our supervision complements existing spatial post-training. Beyond benchmark-level gains, dense trajectory evaluation in Section ~\ref{sec:dense} further shows that \textsc{URUQI}$_{\mathrm{Syn}}$-8B improves self-motion estimation and maintenance of previously observed object locations as the observer moves.

\section{Related Work}
\label{sec:related}

\paragraph{Spatial Intelligence in Vision-Language Models.}

Recent work has improved spatial reasoning in VLMs by scaling spatially grounded instruction data for metric and relational reasoning~\citep{chen2024spatialvlm,cai2026scaling}, or by introducing stronger geometric priors such as depth and reconstructed 3D features~\citep{cheng2024spatialrgpt,wu2026spatial,fan2026vlm}. Another line of work delegates geometric operations to specialized tools, with the VLM serving primarily as a semantic reasoner and planner~\citep{dai2026s,chen2026geometrically}. While these approaches demonstrate the value of richer supervision, geometric representations, and external computation, the resulting progress motivates a fundamental question: can spatial intelligence be learned as a native capability of VLMs, enabling them to build and update spatial understanding from visual experience? We study this question by using privileged geometry to supervise how spatial information evolves during training.

\paragraph{Intermediate Supervision for Spatial Reasoning}
Recent work has complemented final-answer supervision by explicitly supervising intermediate spatial reasoning processes. Spatial CoT supervises linguistic reasoning traces~\citep{li2026star}. 
Cognitive-map methods construct explicit spatial scaffolds for subsequent reasoning or verification~\citep{wang2025mindcube,deng2026active}. 
Representation-based approaches instead supervise intermediate visual or geometric states, such as canonical views or latent 3D representations~\citep{zhan20263viewsense,chen2026think}.
These approaches demonstrate the benefit of exposing intermediate structure during spatial reasoning. However, their intermediate targets are typically defined by a particular reasoning format or spatial representation. \name{} instead derives supervision directly from scene geometry, capturing observer motion and spatial information updates as shared components of spatial reasoning.

\paragraph{Spatial Reasoning from Egocentric Visual Experience}

Recent work has extended spatial reasoning from images to egocentric videos, studying how models integrate spatial evidence across changing viewpoints and long temporal contexts~\citep{yang2025thinking,ravi2025out,yang2026cambrian}. A growing line of work further maintains persistent spatial information through object-centric memories, learned spatial memories, or explicit 3D scene representations~\citep{fan2025embodied,liu2026spatial}. 
From a complementary cognitive perspective, self-motion supports continuous spatial updating without requiring complete scene reconstruction~\citep{wang2000updating,wolbers2008spatial}.
Motivated by human spatial cognition,~\name{} instead learns to maintain spatial understanding across observer motion, supporting persistent object mapping and subsequent spatial reasoning without complete scene reconstruction.
\begin{figure*}[t]
    \centering
    \includegraphics[width=0.9\linewidth]{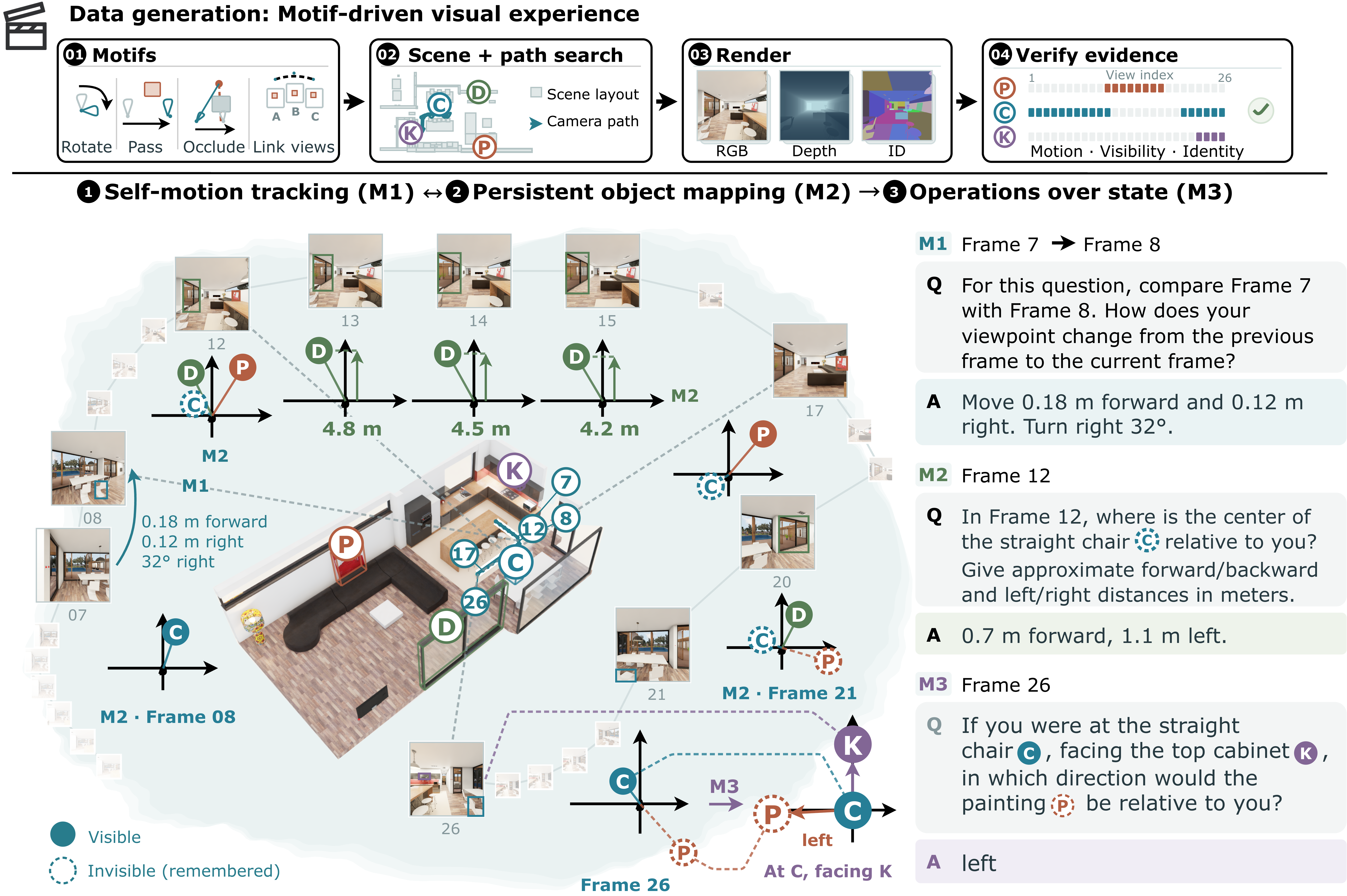}

    \caption{\textbf{Overview of \name{}.}
    \textbf{Top:} Visual experiences are generated by instantiating reusable motifs in 3D scenes, searching for feasible camera trajectories, and rendering RGB observations together with privileged depth and instance identities. 
    \textbf{Bottom:} Each verified trajectory is compiled into supervision over three stages: self-motion tracking (M1), persistent object mapping (M2), and operations over spatial state (M3). The example illustrates spatial information evolving along a shared trajectory, with solid and dashed markers denoting visible and previously observed objects.}
    \label{fig:method}
    \vspace{-2ex}
\end{figure*}

\section{Methodology}
\label{sec:method}

In this section, we first formulate spatial supervision over visual experience in Section~\ref{subsec:problem}. We then introduce motif-driven trajectory rendering for generating visual experiences in Section~\ref{subsec:motif}. Finally, Section~\ref{subsec:compile} describes how privileged scene geometry is used to derive spatial supervision from visual experiences and organize resulting targets into training episodes with shared context.

\subsection{Problem formulation}
\label{subsec:problem}

We consider a static physical world $\mathcal{W}$ observed along a moving camera trajectory $\tau=(T_1,\ldots,T_L)$, where $T_t\in\mathrm{SE}(3)$ denotes the camera pose at step $t$. The trajectory yields an RGB sequence $I_{1:L}$, which constitutes the observer's visual experience. The VLM observes only $I_{1:L}$, while the world geometry $\mathcal{W}$ and camera poses $T_{1:L}$ are privileged information used only to construct supervision.

For two consecutive observations, the relative observer motion is given by:
\begin{equation}
\Delta T_t = T_t^{-1}T_{t+1}\in\mathrm{SE}(3).
\end{equation}
This motion changes the observer's reference frame and therefore constrains how previously acquired spatial information should be interpreted at the next viewpoint. Let $S_t$ denote the supervision-level spatial state supported by observations up to step $t$. Upon receiving $I_{t+1}$, the state must be updated consistently with both the relative motion and the new visual evidence:
\begin{equation}
S_t
\xrightarrow{\Delta T_t,I_{t+1}}
S_{t+1}.
\end{equation}
Here, $S_t$ is an abstraction for supervision and does not assume that the VLM maintains an explicit map or a particular internal state representation. A spatial query $q$ issued at step $t_q$ specifies how the available spatial information should be used:
\begin{equation}
y_q=g_q(S_{t_q}),
\end{equation}
where $g_q$ denotes the query-dependent spatial operation and $y_q$ is its target answer. This factorization exposes three supervision targets: observer motion $\Delta T_t$, spatial-state transitions $S_t \rightarrow S_{t+1}$, and query-dependent operations over the resulting state.

\subsection{Motif-driven Visual Experience Generation}
\label{subsec:motif}
The same physical world can present distinct visual experiences depending on how it is observed. Changes in visual experience reveal different geometric evidence and place distinct demands on spatial reasoning.
For example, rotating in place changes the observer's orientation while keeping its position fixed, isolating changes in the reference frame. Exploring and revisiting a scene requires spatial information to persist across viewpoints and temporary loss of visibility. We aim to design a broad set of visual experiences to evaluate spatial reasoning under different conditions.

As shown in the top of Figure~\ref{fig:method}, we control the generation of visual experience through \emph{spatial motifs}: reusable patterns of observer motion and visibility that determine how spatial evidence unfolds along a trajectory.
Given a 3D scene, we instantiate a motif by selecting the relevant objects and searching for a feasible trajectory under the scene geometry.
Each motif then defines how the observer should move and how spatial information should be revealed, such as rotating at a fixed location, moving until a previously seen object becomes occluded, or walking through landmarks. 

The motif is instantiated through constrained trajectory search over the scene geometry, jointly determining camera positions, motion paths, and viewing directions.
Detailed motif definitions and the corresponding search procedures are provided in Appendix~\ref{app:motif_instantiation}.
Each candidate trajectory is then executed in OmniGibson~\citep{li2022behavior} to produce the RGB sequence together with privileged geometric signals, including camera poses, depth, and instance identities. We then verify that these observations and signals satisfy the
motion, visibility, and object-identity requirements of the motif.

Using this procedure, we construct 11,738 visual trajectories, covering a broad range of ways in which spatial information can emerge, disappear, and reconnect over time. Detailed trajectory statistics are provided in Appendix~\ref{app:trajectory_statistics}.

\subsection{Compiling Spatial Supervision into Episodes}
\label{subsec:compile}

As shown in the bottom of Figure~\ref{fig:method}, for each visual trajectory, we use its privileged camera poses, object identities, and visibility annotations to instantiate the spatial supervision defined in Section~\ref{subsec:problem}. We first derive supervision targets from geometric quantities and then express them as natural-language QA pairs. These targets are subsequently placed along the trajectory according to the visual evidence available at each point to form a temporally grounded training episode.

\textbf{Geometric supervision.} The trajectory geometry provides three complementary forms of supervision. For \textbf{self-motion tracking} (M1), relative camera poses determine the translation and heading change between viewpoints, including both consecutive and longer-range motion along the trajectory. For \textbf{persistent object mapping} (M2), object identities associate entities across observations, while object geometry and camera poses determine their locations and relations in the reference frame required by each query. These targets cover both currently visible objects and previously observed objects that have moved outside the current field of view. For \textbf{operations over state} (M3), the query-dependent operator \(g_q\) acts on the spatial information represented by \(S_t\) to produce targets such as reference-frame transformations, hypothetical motion updates, metric or directional comparisons, and compositions of information acquired across observations.

\textbf{Temporal grounding.} We align each supervision target with the point in the trajectory at which its supporting visual evidence becomes available. At step \(t\), a target may involve only entities and spatial information established by observations \(I_{1:t}\). Once an entity has been observed, its spatial information remains available for subsequent targets as the observer moves, including after the entity becomes occluded or leaves the current view. Entities not yet encountered are introduced only after their first supporting observation. Accordingly, self-motion targets follow the observations defining the corresponding relative pose, object-mapping targets follow the observations establishing the relevant entities and spatial information, and operations targets are introduced once their required spatial information and reference frame are available.

\textbf{Episode construction.}
Let \(t_1 < \cdots < t_K\) denote the points at which the \(K\) supervision queries are inserted. We serialize the trajectory and its associated question--answer pairs into a single episode:
$$
\mathcal{E}
=
\big(
I_{1:t_1}, q_1, y_1;\;
I_{t_1+1:t_2}, q_2, y_2;\;
\ldots;\;
I_{t_{K-1}+1:t_K}, q_K, y_K
\big).
$$
Observations retain their temporal order, while the context accumulates as the episode progresses. Thus, when answering \(q_k\), the model has access to preceding visual observations and earlier QA exchanges. M1, M2, and M3 targets are interleaved along the visual trajectory, placing different forms of spatial supervision within the same evolving context.

\paragraph{Training objective.}
We train ~\name{} models using autoregressive next-token prediction on answer tokens.
Let $i$ index the $N$ QA occurrences across all training episodes.
For QA $i$, the context $H_i$ contains the available images, the current question,
and any preceding QA exchanges in the same episode.
Its target answer is the token sequence $y_i=(y_{i,1},\ldots,y_{i,n_i})$,
where $n_i$ is the number of answer tokens.
The training objective is:
\begin{equation}
\mathcal{L}(\theta)
= -\frac{1}{N}\sum_{i=1}^{N}\frac{1}{n_i}
\sum_{j=1}^{n_i}
\log p_{\theta}\bigl(y_{i,j}\mid H_i,y_{i,<j}\bigr),
\end{equation}
where $y_{i,<j}$ denotes the target answer tokens preceding position $j$.

\section{Experiments}
\label{sec:experiments}
In this section, we conduct experiments to address the following research questions:
\begin{packeditemize}
    \item  \textbf{RQ1}: Does post-training on continuous visual experience improve the underlying spatial capabilities targeted by our supervision?  
    \item \textbf{RQ2}: Do these gains transfer to external spatial benchmarks and real-world visual observations?
    \item \textbf{RQ3}: Can VLMs maintain egocentric spatial understanding as the observer moves?
\end{packeditemize}

\begin{table}[t]
\centering
\caption{\textbf{Evaluation on \name{} benchmark.} Each column reports task-specific accuracy (\%). Subclasses are grouped by the three supervision stages. Overall is computed as the mean of the three stage scores, with subclasses weighted equally within each stage. Bold and underline denote the best and second-best scores, respectively.}
\label{tab:indomain}
\begingroup
\footnotesize
\setlength{\tabcolsep}{2.8pt}
\renewcommand{\arraystretch}{1.10}
\begin{tabularx}{\linewidth}{@{}>{\raggedright\arraybackslash}X*{12}{r}@{}}
\toprule
Model & \multicolumn{3}{c}{M1: self-motion}
      & \multicolumn{4}{c}{M2: object mapping}
      & \multicolumn{4}{c}{M3: state operations}
      & \textbf{Overall} \\
\cmidrule(lr){2-4}\cmidrule(lr){5-8}\cmidrule(lr){9-12}
 & Mot. & Loc. & Yaw & ID & Pos. & Hist. & Inv.
 & Ref. & Upd. & Rel. & Evt. & $\uparrow$ \\

\midrule
\multicolumn{13}{l}{\textit{Proprietary models}} \\
Gemini 3.1 Pro & 13.64 & 17.96 & 43.58 & 29.41 & 11.84 & 25.15 & 48.65 & 25.40 & 34.82 & 27.37 & 25.56 & 27.37 \\
Qwen3.8-Max & 9.77 & 16.45 & 39.57 & \underline{32.10} & 12.85 & 24.08 & 47.27 & 14.75 & 20.01 & 24.58 & 19.60 & 23.58 \\
GPT-5.5 & 11.86 & 21.16 & 44.53 & 31.22 & 11.36 & 20.18 & 44.20 & 22.43 & 19.04 & 21.30 & 20.90 & 24.50 \\
GPT-6 Astra & 25.31 & 41.30 & \textbf{89.70} & \textbf{52.53} & \underline{29.72} & \textbf{54.91} & \textbf{61.38} & 47.13 & \textbf{63.25} & \underline{45.89} & 37.68 & \underline{50.08} \\

\midrule
\multicolumn{13}{l}{\textit{General-purpose open-source models}} \\
InternVL3-8B & 5.47 & 22.40 & 29.18 & 8.08 & 9.71 & 7.03 & 26.66 & 12.61 & 11.18 & 21.10 & 17.62 & 15.84 \\
Qwen3-VL-8B & 7.49 & 17.60 & 30.17 & 13.02 & 11.01 & 10.92 & 35.89 & 16.17 & 11.55 & 23.75 & 27.32 & 18.61 \\
Qwen3.8-27B & 7.80 & 13.48 & 33.75 & 19.83 & 9.37 & 15.01 & 37.76 & 9.15 & 14.90 & 16.53 & 7.30 & 16.94 \\

\midrule
\multicolumn{13}{l}{\textit{Spatially specialized open-source models}} \\
SenseNova-SI-1.5 & 2.09 & 12.99 & 35.34 & 6.81 & 10.82 & 10.87 & 24.14 & 22.92 & 11.25 & 24.71 & 25.47 & 17.02 \\
Spatial-MLLM & 0.00 & 1.50 & 6.85 & 0.09 & 2.93 & 1.51 & 3.93 & 5.99 & 4.94 & 11.13 & 18.25 & 4.99 \\
SpaceR-7B & 0.18 & 6.95 & 30.47 & 3.36 & 10.23 & 6.56 & 21.38 & 12.67 & 10.26 & 17.86 & 31.18 & 13.64 \\
VST-7B-SFT & 0.00 & 5.19 & 28.02 & 2.46 & 9.79 & 10.14 & 13.40 & 16.35 & 10.48 & 17.86 & 31.14 & 12.99 \\
MindCube-RL & 0.00 & 4.93 & 29.44 & 2.00 & 8.17 & 5.17 & 10.80 & 10.75 & 10.19 & 9.47 & 22.03 & 10.37 \\

\midrule
\multicolumn{13}{l}{\textit{URUQI family}} \\
\rowcolor{black!5}
\textsc{URUQI}$_{\text{Syn}}$-8B & \textbf{33.48} & \underline{54.91} & 71.08 & 24.07 & 28.03 & 37.68 & 52.82 & 59.49 & 44.08 & 41.07 & 72.93 & 47.73 \\
\rowcolor{black!5}
\textsc{URUQI}-Mix-8B & \underline{33.05} & 54.88 & 69.27 & 25.49 & 27.85 & 39.05 & \underline{53.19} & \underline{60.19} & 43.56 & 44.60 & \underline{76.63} & 48.35 \\
\rowcolor{black!5}
\textsc{URUQI}-SI-Mix-8B & 30.10 & \textbf{56.45} & \underline{71.68} & 25.84 & \textbf{30.27} & \underline{42.13} & 51.38 & \textbf{66.31} & \underline{47.25} & \textbf{46.47} & \textbf{84.31} & \textbf{50.41} \\
\bottomrule
\end{tabularx}
\endgroup
\end{table}

\begin{table}[t]
\centering
\caption{\textbf{External spatial evaluation.}
Num and MC denote numerical and multiple-choice tasks in VSI-Bench;
Pos and MSR denote positional and multi-step spatial reasoning in MMSI-Bench;
Rot, Amg, and Ard denote rotation, among, and around in MindCube-tiny.
Avg reports each benchmark's overall score.
Bold and underline indicate the best and second-best scores among open-source models, respectively.
Superscripts indicate externally reported results:
\textsuperscript{a}SpatialAxiom~\citep{spatialaxiom};
\textsuperscript{b}Wang~\citep{wang2026astra};
\textsuperscript{c}PhysBrain~1.5~\citep{physbrain15}
.}
\label{tab:external}
\begingroup
\footnotesize
\setlength{\tabcolsep}{4pt}
\renewcommand{\arraystretch}{1.10}
\providecommand{\na}{--}
\begin{tabularx}{\columnwidth}{@{}>{\raggedright\arraybackslash}X*{10}{r}@{}}
\toprule
& \multicolumn{3}{c}{VSI-Bench}
& \multicolumn{3}{c}{MMSI-Bench}
& \multicolumn{4}{c}{MindCube-tiny} \\
\cmidrule(lr){2-4}\cmidrule(lr){5-7}\cmidrule(l){8-11}
Model & Num & MC & \textbf{Avg} & Pos. & MSR & \textbf{Avg}
      & Rot & Amg & Ard & \textbf{Avg} \\
\midrule
\multicolumn{11}{@{}l}{\textit{Proprietary models (reference)}} \\
Gemini 3.1 Pro & 38.48 & 61.36 & 49.92 & \na & \na & 49.50\textsuperscript{a} & 90.50 & 71.33 & 83.20 & 77.81 \\
Qwen3.7-Plus & 61.56 & 68.51 & 65.04 & \na & \na & 45.00\textsuperscript{a} & 92.50 & 59.83 & 80.40 & 70.95 \\
GPT-5.5 & \na & \na & 60.40\textsuperscript{a} & \na & \na & 42.20\textsuperscript{a} & \na & \na & \na & 65.50\textsuperscript{a} \\
GPT-6 Astra & 64.57 & 81.53 & 73.05\textsuperscript{b} & \na & \na & 57.90\textsuperscript{c} & \na & \na & \na & 78.80\textsuperscript{c} \\

\midrule
\multicolumn{11}{@{}l}{\textit{General-purpose open-source models}} \\
InternVL3-8B & 48.02 & 36.26 & 42.14 & 30.84 & 21.72 & 27.80 & 33.00 & 34.33 & 51.60 & 38.19 \\
Qwen3-VL-8B & 63.19 & 53.32 & 58.25 & 30.84 & 28.28 & 29.40 & 29.50 & 28.33 & 32.00 & 29.43 \\
Qwen3.8-27B & 48.40 & 53.10 & 50.75 & 40.80 & \textbf{37.88} & \textbf{42.70} & \textbf{92.50} & \underline{67.33} & 83.60 & 76.00 \\

\midrule
\multicolumn{11}{@{}l}{\textit{Spatially specialized open-source models}} \\
\shortstack[l]{SenseNova-SI-1.5}
  & \underline{65.85} & \underline{68.57} & \underline{67.21} & \underline{45.40} & 28.28 & 39.10 & \underline{90.50} & \textbf{93.83} & \underline{88.80} & \underline{92.00} \\
\shortstack[l]{Spatial-MLLM}
  & 50.96 & 41.49 & 46.33 & 27.16 & \underline{29.80} & 26.10 & 39.00 & 30.51 & 36.00 & 33.46 \\
SpaceR-7B & 49.85 & 41.55 & 45.60 & 35.29 & 23.23 & 27.80 & 34.50 & 31.02 & 33.60 & 32.31 \\
VST-7B-SFT & 60.72 & 50.28 & 55.50 & 35.35 & 18.18 & 32.50 & 37.00 & 35.93 & 50.80 & 39.71 \\
VST-7B-RL & 61.21 & 51.10 & 56.15 & 36.13 & 19.19 & 32.50 & 37.00 & 37.80 & 45.60 & 39.52 \\
\shortstack[l]{MindCube-3B-SFT}
  & 15.87 & 18.62 & 17.24 & 2.15 & 2.02 & 1.70 & 34.00 & 51.02 & 67.60 & 51.73 \\
MindCube-RL & 25.85 & 37.09 & 31.47 & 27.39 & 27.78 & 27.70 & 30.50 & 51.17 & 61.60 & 49.71 \\
Cambrian-S-7B & 63.52 & 62.34 & 62.92 & 29.56 & 24.24 & 27.10 & 33.00 & 38.98 & 39.20 & 37.88 \\

\midrule
\multicolumn{11}{@{}l}{\textit{URUQI family}} \\
\rowcolor{black!5}
\shortstack[l]{\textsc{URUQI}$_{\mathrm{Syn}}$-8B}
  & 39.28 & 55.23 & 47.26 & 34.67 & 22.22 & 33.10 & 44.00 & 50.33 & 36.80 & 45.90 \\
\rowcolor{black!5}
\shortstack[l]{\textsc{URUQI}-Mix-8B}
  & 55.38 & 59.73 & 57.56 & 45.02 & 21.72 & 39.80 & 49.00 & 57.83 & 57.20 & 56.00 \\
\rowcolor{black!5}
\shortstack[l]{\textsc{URUQI}-SI-Mix-8B}
  & \textbf{66.39} & \textbf{69.12} & \textbf{67.76} & \textbf{50.19} & 28.79 & \underline{41.80} & 90.00 & \textbf{93.83} & \textbf{91.60} & \textbf{92.57} \\
\bottomrule
\end{tabularx}

\endgroup
\end{table}

\subsection{Experimental Setup}
\label{subsec:setup}
\textbf{Training data and evaluation data.}
The \name{} training corpus contains 611,348 QA instances
in 455,863 episodes: 211,552 for self-motion tracking,
136,398 for persistent object mapping, and 263,398 for
spatial-state operations.
The benchmark contains 52,920 QA instances in 2,692 episodes
from eight held-out scenes
(Appendix~\ref{app:trajectory_statistics}).

\textbf{Training setting.}
We post-train all models in the \name{} family for one epoch on 16 NVIDIA H800 GPUs using AdamW with a learning rate of $10^{-5}$ and an effective batch size of 192 training episodes. We consider three URUQI-8B configurations.
\begin{packeditemize}
    \item \textbf{\textsc{URUQI}$_{\text{Syn}}$-8B}: initialized from InternVL3-8B~\citep{zhu2025internvl3} and post-trained only on our synthesized visual-experience supervision.
    \item \textbf{\textsc{URUQI}-Mix-8B}: initialized from
    InternVL3-8B and trained on a mixture of our visual-experience supervision and 100k examples randomly selected from SenseNova-SI-8M~\citep{cai2026scaling}. 
    \item \textbf{\textsc{URUQI}-SI-Mix-8B}: initialized from
    SenseNova-SI-1.5-InternVL3-8B~\citep{cai2026scaling} and further trained with the same mixed-data recipe as \textsc{URUQI}-MIX-8B.
\end{packeditemize}

\textbf{Evaluation setting}
We evaluate spatial capabilities on \name{} benchmark in a multi-turn setting in Table~\ref{tab:indomain}.
Observations and questions are presented in temporal order, and the model retains preceding observations and its own answers within each episode.
We further evaluate models on VSI-Bench~\citep{yang2025thinking}, MMSI-Bench~\citep{yang2026mmsi}, and MindCube-tiny~\citep{wang2025mindcube} using their respective evaluation protocols in Table~\ref{tab:external}. 
We compare ~\name{} model family against proprietary VLMs including Gemini 3.1 Pro~\citep{googledeepmind2026gemini31pro},
Qwen3.7-Plus~\citep{qwen2026qwen37plus}, Qwen3.8-Max~\citep{qwen38}, GPT-5.5~\citep{openai2026gpt55},
and GPT-6 Astra~\citep{openai2026gpt6astra}, general-purpose open-source VLMs including
InternVL3-8B~\citep{zhu2025internvl3},
Qwen3-VL-8B~\citep{bai2025qwen3vltechnicalreport},
and Qwen3.8-27B~\citep{qwen2026qwen38_27b};
and spatially specialized models including
SenseNova-SI-1.5~\citep{cai2026scaling},
Spatial-MLLM~\citep{wu2026spatial},
SpaceR-7B~\citep{ouyang2025spacer},
VST-7B-SFT and VST-7B-RL~\citep{yang2026visual},
MindCube-3B-SFT and MindCube-RL~\citep{wang2025mindcube},
and Cambrian-S-7B~\citep{yang2026cambrian}.

\subsection{Evaluating Spatial Capabilities in Continuous Visual Experience (RQ1)}
\label{sec:indomain}
\textbf{Metrics.}
Table~\ref{tab:indomain} reports three groups of task-specific accuracies.
M1 measures inter-view translation and rotation (Mot.), relative observer localization (Loc.), and heading changes (Yaw).
M2 measures cross-view correspondence and referent resolution (ID), object positions, distances, and directions (Pos.), historical spatial states and visibility (Hist.), and distinct-object counts across views (Inv.).
M3 measures reference-frame transformations (Ref.), hypothetical and composed motion updates (Upd.), metric relations and comparisons (Rel.), and counts of observed turns (Evt.).
Overall averages the three stage scores, each computed as the mean of its subclasses.

\textbf{Dense experience supervision improves the targeted capabilities.}
Table~\ref{tab:indomain} evaluates self-motion estimation (M1),
object mapping (M2), and operations over spatial state (M3).
Post-training InternVL3-8B solely on our synthetic supervision
raises its Overall score from 15.84\% to 47.73\%, with improvements in all 11 subclasses.

\textbf{Supervision from experience complements existing spatial post-training.}
\textsc{URUQI}-Mix-8B reaches 48.35\% Overall, compared with
47.73\% for \textsc{URUQI}$_{\mathrm{Syn}}$-8B.
Starting from SenseNova-SI-1.5, \textsc{URUQI}-SI-Mix-8B increases Overall from 17.02\% to 50.41\%, improving all 11 subclasses.
It also obtains the highest Overall score
among the evaluated models, closely followed by GPT-6 Astra
at 50.08\%. GPT-6 Astra retains a higher M2 average
(49.64\% vs 37.41\%).

\subsection{Evaluating Transfer to External Spatial Benchmarks (RQ2)}
\label{sec:exp_transfer}

\textbf{Benchmarks and metrics.}
We evaluate transfer on three external spatial benchmarks:
VSI-Bench~\citep{yang2025thinking}, MMSI-Bench~\citep{yang2026mmsi}, and MindCube-tiny~\citep{wang2025mindcube}.
Table~\ref{tab:external} reports their overall scores together
with task-group results.

\textbf{Synthetic supervision transfers to external spatial tasks.}
Compared with InternVL3-8B, \textsc{URUQI}$_{\mathrm{Syn}}$-8B
improves the overall scores from 42.14\% to 47.26\% on VSI-Bench,
from 27.80\% to 33.10\% on MMSI-Bench, and from 38.19\% to
45.90\% on MindCube-tiny.
These gains show that the benefits of our synthetic supervision
extend beyond the tasks in \name{} benchmark.

\textbf{Combining synthetic and existing spatial data improves transfer.}
With the addition of 100K examples from SenseNova-SI-8M (1.25\% of the corpus),
\textsc{URUQI}-Mix-8B reaches 57.56\%, 39.80\%, and 56.00\%
on VSI-Bench, MMSI-Bench, and MindCube-tiny, respectively,
improving over the synthetic-only variant on all three benchmarks.
VSI-Bench Num also increases to 55.38\%, exceeding the
InternVL3-8B baseline.
Starting from SenseNova-SI-1.5,
\textsc{URUQI}-SI-Mix-8B achieves the highest overall scores
on VSI-Bench and MindCube-tiny among the open-source models
in Table~\ref{tab:external}, while ranking second on
MMSI-Bench, behind only Qwen3.8-27B.

\begin{table*}[t]
\centering
\caption{
\textbf{Dense evaluation of self-motion estimation and egocentric
object localization.}
Acc@0.5m measures the percentage of object predictions
within 0.5\,m of the ground-truth horizontal position.
Bold and underline denote the best and second-best results per column, respectively.
}
\label{tab:atomic-spatial}
\begingroup
\small
\setlength{\tabcolsep}{3.5pt}
\renewcommand{\arraystretch}{1.12}

\begin{tabular*}{\linewidth}{
@{\extracolsep{\fill}}llrrrrrrr@{}
}
\toprule
\multirow{3}{*}{Model}
& \multirow{3}{*}{QA history}
& \multicolumn{2}{c}{Self-motion}
& \multicolumn{5}{c}{Object mapping} \\
\cmidrule(lr){3-4}\cmidrule(lr){5-9}
& & \multirow{2}{*}{\shortstack{Translation\\error (m) $\downarrow$}}
  & \multirow{2}{*}{\shortstack{Heading\\error ($^\circ$) $\downarrow$}}
  & \multicolumn{4}{c}{Acc@0.5m (\%) $\uparrow$}
  & \multirow{2}{*}{\shortstack{Mean error\\(m) $\downarrow$}} \\
\cmidrule(lr){5-8}
& & & & All & Initial & Visible & Absent & \\
\midrule

\multirow{3}{*}{InternVL3-8B}
& None
& 0.397 & 93.57
& 1.00 & 0.00 & 2.42 & 0.00
& 3.004 \\
& Model
& 0.341 & 16.57
& 0.67 & 0.00 & 1.45 & 0.13
& 3.040 \\
& GT motion
& 0.189 & 12.72
& 0.78 & 0.00 & 1.82 & 0.09
& 3.052 \\
\midrule

\multirow{3}{*}{Qwen3.8-27B}
& None
& 0.297 & 9.43
& 6.44 & 16.02 & 12.30 & 1.29
& 4.725 \\
& Model
& 0.356 & 7.22
& 9.41 & 22.08 & 17.99 & 1.88
& 3.594 \\
& GT motion
& 0.127 & 4.84
& 8.75 & 23.81 & 16.41 & 1.79
& 3.263 \\
\midrule

\multirow{3}{*}{\textsc{URUQI}$_{\mathrm{Syn}}$-8B}
& None
& 0.154 & 2.36
& \textbf{49.56} & \textbf{75.32} & \textbf{66.14} & \underline{35.13}
& \textbf{0.743} \\
& Model
& \underline{0.090} & \underline{0.97}
& 39.62 & \underline{74.89} & 47.24 & 30.94
& 1.027 \\
& GT motion
& \textbf{0.031} & \textbf{0.56}
& \underline{47.04} & 73.16 & \underline{55.97} & \textbf{38.30}
& \underline{0.832} \\
\bottomrule
\end{tabular*}
\endgroup
\end{table*}

\begin{table*}[t]
\centering
\caption{\textbf{Localization from visual estimates and camera poses.}
Acc@0.5m (\%) is averaged over 2,240 absent-object queries.
(a) Direct prediction uses each current VLM answer; pose-based propagation stores the first accepted
object estimate and computes subsequent locations from camera poses.
Both use No QA history, with GT poses supplied only to the geometry module.
(b) Pose-based propagation uses Model-history object predictions; Predicted denotes
poses integrated from the model's motion estimates.
Bold marks the best result within each model and panel.
Complete metrics are in Table~\ref{tab:pose-assisted-full}.}
\label{tab:pose-assisted-unified}
\begingroup
\small
\setlength{\tabcolsep}{2pt}
\renewcommand{\arraystretch}{1.12}
\begin{minipage}[t]{0.505\linewidth}
\centering
\textbf{(a) Localization method}\\[2pt]
\emph{No QA history; GT poses}\\[4pt]
\begin{tabular*}{\linewidth}{@{\extracolsep{\fill}}llr@{}}
\toprule
Model & Method & \shortstack{Absent\\Acc@0.5m $\uparrow$} \\
\midrule
\multirow{2}{*}{Qwen3.8-27B}
& Direct prediction & 1.29 \\
& Pose-based propagation & \textbf{16.88} \\
\midrule
\multirow{2}{*}{\textsc{URUQI}$_{\mathrm{Syn}}$-8B}
& Direct prediction & 35.13 \\
& Pose-based propagation & \textbf{71.61} \\
\bottomrule
\end{tabular*}
\end{minipage}\hfill
\begin{minipage}[t]{0.465\linewidth}
\centering
\textbf{(b) Camera-pose source}\\[2pt]
\emph{Model history; pose-based propagation}\\[4pt]
\begin{tabular*}{\linewidth}{@{\extracolsep{\fill}}llr@{}}
\toprule
Model & Pose & \shortstack{Absent\\Acc@0.5m $\uparrow$} \\
\midrule
\multirow{2}{*}{Qwen3.8-27B}
& GT & \textbf{18.48} \\
& Predicted & 3.97 \\
\midrule
\multirow{2}{*}{\textsc{URUQI}$_{\mathrm{Syn}}$-8B}
& GT & \textbf{72.86} \\
& Predicted & 34.02 \\
\bottomrule
\end{tabular*}
\end{minipage}
\endgroup
\end{table*}

\subsection{Dense Evaluation of Egocentric Spatial Tracking (RQ3)}
\label{sec:dense}
We evaluate self-motion estimation and egocentric object localization on 100 trajectories in \name{} benchmark, comprising 2,312 frames and 231 registered objects.

To examine whether previous answers help subsequent localization, we compare three history conditions, all of which provide the full observed image history.
\emph{No QA history} excludes previous questions and answers; \emph{Model history} includes the model's earlier motion and location answers; \emph{GT motion history} replaces earlier motion answers with ground truth.

Table~\ref{tab:atomic-spatial} reports motion errors and
localization metrics, including Acc@0.5m, the percentage
of predictions within 0.5\,m of the true horizontal position.
Construction rules, visibility stages, and scoring details
are in Appendix~\ref{app:dense_evaluation}.

\textbf{Spatial tracking from visual history.}
Under No QA history, \textsc{URUQI}$_{\mathrm{Syn}}$-8B
improves both motion estimation and object localization
over InternVL3-8B and Qwen3.8-27B.
Its overall localization accuracy reaches 49.56\%, compared
with 1.00\% and 6.44\%, respectively.
For absent objects, it achieves 35.13\%, versus 0.00\%
and 1.29\%, showing improved localization out of view.

\textbf{Effect of answer history.} For \textsc{URUQI}$_{\mathrm{Syn}}$-8B, Model history improves motion estimation but reduces overall object-localization accuracy from 49.56\% to 39.62\%.
GT motion history partially recovers overall object-localization accuracy and yields the highest absent-object accuracy among the three conditions (38.30\%).
The improvement over Model history suggests that accurate motion history provides useful context for absent-object localization.

\textbf{Pose-assisted localization.}
We test whether an external geometry module can use camera poses
to propagate each object's first accepted location estimate.
Under \emph{No QA history}, supplying GT poses only to this module
improves \textsc{URUQI}$_{\mathrm{Syn}}$-8B's absent-object accuracy
from 35.13\% with direct prediction to 71.61\% with propagation
(Table~\ref{tab:pose-assisted-unified}a).

In a separate comparison, we fix the \emph{Model history} object
predictions and propagation method, replacing GT poses with poses
obtained by integrating model-predicted self-motion.
Accuracy decreases from 72.86\% to 34.02\%, compared with 3.97\%
for Qwen3.8-27B using predicted poses
(Table~\ref{tab:pose-assisted-unified}b).
Thus, the initial object estimates support pose-assisted localization,
but its accuracy remains sensitive to errors in the estimated camera trajectory.

\section{Conclusion}

We study how spatial cognition in VLMs can be learned from continuous visual experience. \name{} couples consecutive observations through observer self-motion and jointly supervises motion tracking, persistent object mapping, and spatial operations. Our 8B model achieves the best overall performance among evaluated open-source models, reaching 50.41\% on \name{} benchmark and leading open-source results on two of three external benchmarks. It matches GPT-6 Astra on \name{} benchmark and surpasses several proprietary models on external benchmarks, highlighting continuous visual experience as an effective, scalable supervision paradigm for spatially capable VLMs.

\subsection*{AI use statement}
Generative AI tools were used to assist with implementing the trajectory
generation and spatial-supervision compilation pipeline, as well as language
polishing, manuscript organization and the presentation of experimental results.  All codes, references, experimental results, and content were reviewed and verified by the authors, who take responsibility for the final text, methods, and claims.

\subsection*{Ethics statement}

Our spatial supervision is generated from simulated scenes and does not involve human participants or personally identifiable information. All experiments are conducted in simulated or benchmark environments, and any released data, models, and derived assets will follow applicable licensing and usage requirements. The resulting models are developed for research purposes and are not evaluated for safety-critical deployment.

\subsection*{Reproducibility statement}
We provide details of data construction, model training, evaluation protocols, and implementation settings in the main text and appendix. The appendix further documents the procedures used for trajectory generation, spatial supervision, and evaluation. Where permitted, we will release the configurations, checkpoints, and evaluation pipeline.

\bibliography{iclr2027_conference}

@inproceedings{chen2024spatialvlm,
  title={Spatialvlm: Endowing vision-language models with spatial reasoning capabilities},
  author={Chen, Boyuan and Xu, Zhuo and Kirmani, Sean and Ichter, Brain and Sadigh, Dorsa and Guibas, Leonidas and Xia, Fei},
  booktitle={Proceedings of the IEEE/CVF conference on computer vision and pattern recognition},
  pages={14455--14465},
  year={2024}
}

@inproceedings{
    li2022behavior,
    title={{BEHAVIOR}-1K: A Benchmark for Embodied {AI} with 1,000 Everyday Activities and Realistic Simulation},
    author={Chengshu Li and Ruohan Zhang and Josiah Wong and Cem Gokmen and Sanjana Srivastava and Roberto Mart{\'\i}n-Mart{\'\i}n and Chen Wang and Gabrael Levine and Michael Lingelbach and Jiankai Sun and Mona Anvari and Minjune Hwang and Manasi Sharma and Arman Aydin and Dhruva Bansal and Samuel Hunter and Kyu-Young Kim and Alan Lou and Caleb R Matthews and Ivan Villa-Renteria and Jerry Huayang Tang and Claire Tang and Fei Xia and Silvio Savarese and Hyowon Gweon and Karen Liu and Jiajun Wu and Li Fei-Fei},
    booktitle={6th Annual Conference on Robot Learning},
    year={2022},
    url={https://openreview.net/forum?id=_8DoIe8G3t}
}

@article{cheng2024spatialrgpt,
  title={Spatialrgpt: Grounded spatial reasoning in vision-language models},
  author={Cheng, An-Chieh and Yin, Hongxu and Fu, Yang and Guo, Qiushan and Yang, Ruihan and Kautz, Jan and Wang, Xiaolong and Liu, Sifei},
  journal={Advances in Neural Information Processing Systems},
  volume={37},
  pages={135062--135093},
  year={2024}
}

@inproceedings{cai2026scaling,
  title={Scaling spatial intelligence with multimodal foundation models},
  author={Cai, Zhongang and Wang, Ruisi and Gu, Chenyang and Pu, Fanyi and Xu, Junxiang and Wang, Yubo and Yin, Wanqi and Yang, Zhitao and Wei, Chen and Zhou, Tongxi and others},
  booktitle={Proceedings of the IEEE/CVF Conference on Computer Vision and Pattern Recognition},
  pages={7879--7890},
  year={2026}
}

@inproceedings{yang2026mmsi,
  title={Mmsi-bench: A benchmark for multi-image spatial intelligence},
  author={Yang, Sihan and Xu, Runsen and Xie, Yiman and Yang, Sizhe and Li, Mo and Lin, Jingli and Zhu, Chenming and Chen, Xiaochen and Duan, Haodong and Yue, Xiangyu and others},
  booktitle={International Conference on Learning Representations},
  volume={2026},
  pages={157051--157088},
  year={2026}
}

@misc{openai2026gpt55,
  author       = {{OpenAI}},
  title        = {{GPT-5.5 System Card}},
  year         = {2026},
  howpublished = {\url{https://openai.com/index/gpt-5-5-system-card/}}
}

@inproceedings{spatialcot,
    title     = {SpatialCoT: Advancing Spatial Reasoning through Coordinate Alignment and Chain-of-Thought for Embodied Task Planning},
    author    = {Yuecheng Liu and Dafeng Chi and Shiguang Wu and Zhanguang Zhang and Yaochen Hu and Lingfeng Zhang and Yingxue Zhang and Shuang Wu and Tongtong Cao and Guowei Huang and Helong Huang and Guangjian Tian and Weichao Qiu and Quan and Jianye Hao and Yuzheng Zhuang},
    year      = {2025},
    booktitle = {arxiv 2025}
}

@article{deng2026active,
  title={Active exploring like a pigeon: Reinforcing spatial reasoning via agentic vision-language models},
  author={Deng, Wei and Zhang, Xianlin and Qi, Mengshi},
  journal={arXiv preprint arXiv:2606.02459},
  year={2026}
}

@misc{wang2025mindcube,
      title={MindCube: Spatial Mental Modeling from Limited Views},
      author={Qineng Wang and Baiqiao Yin and Pingyue Zhang and Jianshu Zhang and Kangrui Wang and Zihan Wang and Jieyu Zhang and Keshigeyan Chandrasegaran and Han Liu and Ranjay Krishna and Saining Xie and Jiajun Wu and Li Fei-Fei and Manling Li},
      year={2025},
      eprint={2506.21458},
      archivePrefix={arXiv},
      primaryClass={cs.AI},
      url={https://arxiv.org/abs/2506.21458},
}

@inproceedings{li2026spatialladder,
  title={Spatialladder: Progressive training for spatial reasoning in vision-language models},
  author={Li, Hongxing and Li, Dingming and Wang, Zixuan and Yan, Yuchen and Wu, Hang and Zhang, Wenqi and Shen, Yongliang and Lu, Weiming and Xiao, Jun and Zhuang, Yueting},
  booktitle={International Conference on Learning Representations},
  volume={2026},
  pages={76566--76592},
  year={2026}
}

@inproceedings{yang2025thinking,
  title={Thinking in space: How multimodal large language models see, remember, and recall spaces},
  author={Yang, Jihan and Yang, Shusheng and Gupta, Anjali W and Han, Rilyn and Fei-Fei, Li and Xie, Saining},
  booktitle={2025 IEEE/CVF Conference on Computer Vision and Pattern Recognition (CVPR)},
  pages={10632--10643},
  year={2025},
  organization={IEEE}
}

@inproceedings{hua2026unleashing,
  title={Unleashing spatial reasoning in multimodal large language models via textual representation guided reasoning},
  author={Hua, Jiacheng and Yin, Yishu and Wu, Yuhang and Wang, Tai and Huang, Yifei and Liu, Miao},
  booktitle={Proceedings of the 64th Annual Meeting of the Association for Computational Linguistics (Volume 1: Long Papers)},
  pages={13616--13637},
  year={2026}
}

@inproceedings{marsili2025visual,
  title={Visual agentic ai for spatial reasoning with a dynamic api},
  author={Marsili, Damiano and Agrawal, Rohun and Yue, Yisong and Gkioxari, Georgia},
  booktitle={2025 IEEE/CVF Conference on Computer Vision and Pattern Recognition (CVPR)},
  pages={19446--19455},
  year={2025},
  organization={IEEE}
}

@inproceedings{kancheti2026chain,
  title={Chain-of-Thought Degrades Visual Spatial Reasoning Capabilities of Multimodal LLMs},
  author={Kancheti, Sai Srinivas and Kanade, Aditya Sanjiv and Balasubramanian, Vineeth N and Ganu, Tanuja},
  booktitle={Proceedings of the 64th Annual Meeting of the Association for Computational Linguistics (Volume 2: Short Papers)},
  pages={862--876},
  year={2026}
}

@article{wang2000updating,
  title={Updating egocentric representations in human navigation},
  author={Wang, Ranxiao Frances and Spelke, Elizabeth S},
  journal={Cognition},
  volume={77},
  number={3},
  pages={215--250},
  year={2000},
  publisher={Elsevier}
}

@article{wolbers2008spatial,
  title={Spatial updating: how the brain keeps track of changing object locations during observer motion},
  author={Wolbers, Thomas and Hegarty, Mary and B{\"u}chel, Christian and Loomis, Jack M},
  journal={Nature neuroscience},
  volume={11},
  number={10},
  pages={1223--1230},
  year={2008},
  publisher={Nature Publishing Group US New York}
}

@article{jiang2024mantis,
  title={Mantis: Interleaved multi-image instruction tuning},
  author={Jiang, Dongfu and He, Xuan and Zeng, Huaye and Wei, Cong and Ku, Max and Liu, Qian and Chen, Wenhu},
  journal={arXiv preprint arXiv:2405.01483},
  year={2024}
}

@article{wu2026spatial,
  title={Spatial-mllm: Boosting mllm capabilities in visual-based spatial intelligence},
  author={Wu, Diankun and Liu, Fangfu and Hung, Yi-Hsin and Duan, Yueqi},
  journal={Advances in neural information processing systems},
  volume={38},
  pages={13569--13597},
  year={2026}
}

@inproceedings{fan2026vlm,
  title={Vlm-3r: Vision-language models augmented with instruction-aligned 3d reconstruction},
  author={Fan, Zhiwen and Zhang, Jian and Li, Renjie and Zhang, Junge and Chen, Runjin and Hu, Hezhen and Wang, Kevin and Wang, Peihao and Qu, Huaizhi and Zhou, Shijie and others},
  booktitle={Proceedings of the IEEE/CVF Conference on Computer Vision and Pattern Recognition},
  pages={31054--31065},
  year={2026}
}

@article{dai2026s,
  title={S-Agent: Spatial Tool-Use Elicits Reasoning for Spatial Intelligence},
  author={Dai, Yalun and Li, Hao and Tian, Shulin and Yao, Runmao and Dong, Yuhao and Hong, Fangzhou and Chen, Zhaoxi and Liu, Fangfu and Tian, Baoliang and Zhang, Dingwen and others},
  journal={arXiv preprint arXiv:2606.20515},
  year={2026}
}

@inproceedings{chen2026geometrically,
  title={Geometrically-constrained agent for spatial reasoning},
  author={Chen, Zeren and Lu, Xiaoya and Zheng, Zhijie and Li, Pengrui and He, Lehan and Zhou, Yijin and Shao, Jing and Zhuang, Bohan and Sheng, Lu},
  booktitle={Proceedings of the IEEE/CVF Conference on Computer Vision and Pattern Recognition},
  pages={38689--38699},
  year={2026}
}

@inproceedings{li2026star,
  title={STAR-R1: Multi-View Spatial TrAnsformation Reasoning by Reinforcing Multimodal LLMs},
  author={Li, Zongzhao and Ma, Zongyang and Li, Mingze and Li, Songyou and Rong, Yu and Xu, Tingyang and Zhang, Ziqi and Zhao, Deli and Huang, Wenbing},
  booktitle={Proceedings of the IEEE/CVF Conference on Computer Vision and Pattern Recognition},
  pages={12041--12051},
  year={2026}
}

@article{zhan20263viewsense,
  title={3viewsense: Spatial and mental perspective reasoning from orthographic views in vision-language models},
  author={Zhan, Shaoxiong and Lai, Yanlin and Liu, Zheng and Lin, Hai and Li, Shen and Cai, Xiaodong and Lin, Zijian and Huang, Wen and Zheng, Hai-Tao},
  journal={arXiv preprint arXiv:2603.07751},
  year={2026}
}

@inproceedings{chen2026think,
  title={Think with 3d: Geometric imagination grounded spatial reasoning from limited views},
  author={Chen, Zhangquan and Zhang, Manyuan and Yu, Xinlei and Luo, Xufang and Sun, Mingze and Pan, Zihao and An, Xiang and Feng, Yan and Pei, Peng and Cai, Xunliang and others},
  booktitle={Proceedings of the IEEE/CVF Conference on Computer Vision and Pattern Recognition},
  pages={2613--2624},
  year={2026}
}

@inproceedings{ravi2025out,
  title={Out of sight, not out of context? egocentric spatial reasoning in vlms across disjoint frames},
  author={Ravi, Sahithya and Sarch, Gabriel Herbert and Vineet, Vibhav and Wilson, Andrew D and Kumaravel, Balasaravanan Thoravi},
  booktitle={Proceedings of the 2025 Conference on Empirical Methods in Natural Language Processing},
  pages={16146--16161},
  year={2025}
}

@inproceedings{yang2026cambrian,
  title={Cambrian-s: Towards spatial supersensing in video},
  author={Yang, Shusheng and Yang, Jihan and Huang, Pinzhi and Brown, Ellis and Yang, Zihao and Yu, Yue and Tong, Shengbang and Zheng, Zihan and Xu, Yifan and Wang, Muhan and others},
  booktitle={International Conference on Learning Representations},
  volume={2026},
  pages={78185--78225},
  year={2026}
}

@inproceedings{fan2025embodied,
  title={Embodied videoagent: Persistent memory from egocentric videos and embodied sensors enables dynamic scene understanding},
  author={Fan, Yue and Ma, Xiaojian and Su, Rongpeng and Guo, Jun and Wu, Rujie and Chen, Xi and Li, Qing},
  booktitle={2025 IEEE/CVF International Conference on Computer Vision (ICCV)},
  pages={6342--6352},
  year={2025},
  organization={IEEE}
}

@inproceedings{liu2026spatial,
  title={Spatial-ttt: Streaming visual-based spatial intelligence with test-time training},
  author={Liu, Fangfu and Wu, Diankun and Chi, Jiawei and Cai, Yimo and Hung, Yi-Hsin and Yu, Xumin and Li, Hao and Hu, Han and Rao, Yongming and Duan, Yueqi},
  booktitle={European Conference on Computer Vision},
  pages={339--357},
  year={2026},
  organization={Springer}
}

@misc{googledeepmind2026gemini31pro,
  author       = {{Google DeepMind}},
  title        = {Gemini 3.1 Pro},
  year         = {2026},
  month        = feb,
  howpublished = {Model Card},
  url          = {https://deepmind.google/models/model-cards/gemini-3-1-pro/},
  note         = {Published February 19, 2026}
}

@misc{qwen2026qwen37plus,
  author       = {{Qwen Team}},
  title        = {Qwen3.7-Plus},
  year         = {2026},
  howpublished = {Model Card},
  url          = {https://www.qwencloud.com/models/qwen3.7-plus}
}

@misc{openai2026gpt6astra,
  author       = {{OpenAI}},
  title        = {{GPT-6 Astra} System Card},
  year         = {2026},
  month        = sep,
  howpublished = {System Card},
  url          = {https://deploymentsafety.openai.com/gpt-6-astra},
  note         = {Published September 3, 2026}
}

@misc{qwen2026qwen38_27b,
  author       = {{Qwen Team}},
  title        = {{Qwen3.8-27B}},
  year         = {2026},
  month        = aug,
  howpublished = {Hugging Face Model Card},
  url          = {https://huggingface.co/Qwen/Qwen3.8-27B},
  note         = {Released August 14, 2026}
}

@article{zhu2025internvl3,
  title={Internvl3: Exploring advanced training and test-time recipes for open-source multimodal models},
  author={Zhu, Jinguo and Wang, Weiyun and Chen, Zhe and Liu, Zhaoyang and Ye, Shenglong and Gu, Lixin and Tian, Hao and Duan, Yuchen and Su, Weijie and Shao, Jie and others},
  journal={arXiv preprint arXiv:2504.10479},
  year={2025}
}

@misc{qwen38,
  title  = {{Qwen3.8-Max}: A New Bar for Coding and Cowork},
  author = {{Qwen Team}},
  year   = {2026},
  month  = {August},
  url    = {https://qwen.ai/blog?id=qwen3.8}
}

@misc{bai2025qwen3vltechnicalreport,
      title={Qwen3-VL Technical Report}, 
      author={Shuai Bai and Yuxuan Cai and Ruizhe Chen and Keqin Chen and Xionghui Chen and Zesen Cheng and Lianghao Deng and Wei Ding and Chang Gao and Chunjiang Ge and Wenbin Ge and Zhifang Guo and Qidong Huang and Jie Huang and Fei Huang and Binyuan Hui and Shutong Jiang and Zhaohai Li and Mingsheng Li and Mei Li and Kaixin Li and Zicheng Lin and Junyang Lin and Xuejing Liu and Jiawei Liu and Chenglong Liu and Yang Liu and Dayiheng Liu and Shixuan Liu and Dunjie Lu and Ruilin Luo and Chenxu Lv and Rui Men and Lingchen Meng and Xuancheng Ren and Xingzhang Ren and Sibo Song and Yuchong Sun and Jun Tang and Jianhong Tu and Jianqiang Wan and Peng Wang and Pengfei Wang and Qiuyue Wang and Yuxuan Wang and Tianbao Xie and Yiheng Xu and Haiyang Xu and Jin Xu and Zhibo Yang and Mingkun Yang and Jianxin Yang and An Yang and Bowen Yu and Fei Zhang and Hang Zhang and Xi Zhang and Bo Zheng and Humen Zhong and Jingren Zhou and Fan Zhou and Jing Zhou and Yuanzhi Zhu and Ke Zhu},
      year={2025},
      eprint={2511.21631},
      archivePrefix={arXiv},
      primaryClass={cs.CV},
      url={https://arxiv.org/abs/2511.21631}, 
}

@article{ouyang2025spacer,
  title={Spacer: Reinforcing mllms in video spatial reasoning},
  author={Ouyang, Kun and Liu, Yuanxin and Wu, Haoning and Liu, Yi and Zhou, Hao and Zhou, Jie and Meng, Fandong and Sun, Xu},
  journal={arXiv preprint arXiv:2504.01805},
  year={2025}
}

@inproceedings{yang2026visual,
  title={Visual spatial tuning},
  author={Yang, Rui and Zhu, Ziyu and Li, Yanwei and Huang, Jingjia and Yan, Shen and Zhou, Siyuan and Liu, Zhe and Li, Xiangtai and Li, Shuangye and Wang, Wenqian and others},
  booktitle={European Conference on Computer Vision},
  pages={192--211},
  year={2026},
  organization={Springer}
}

@misc{wang2026astra,
  author       = {Wang, Yipeng},
  title        = {How Well Does Astra Understand Real-World Space?},
  year         = {2026},
  month        = sep,
  howpublished = {\url{https://www.yipeng.dev/blog/astra-spatial-intelligence}},
  note         = {Independent evaluation. Published September 22, 2026.
                  Accessed September 26, 2026}
}

@misc{spatialaxiom,
    title  = {SpatialAxiom: An Open Spatial Intelligence Model for General Spatial Reasoning},
    author = {Lou, Yujing and Chen, Pingyi and Cao, Shen and Gu, Jiaqi and Guo, Jinhui and Tong, Jintao and Hao, Yunzhuo and Liu, Yao and Wu, Yue and Fan, Lubin and Ye, Jieping},
    month  = {July},
    year   = {2026},
    url    = {https://d2i-ai.github.io/SpatialAxiom}
}

@misc{physbrain15,
      title={PhysBrain 1.5: From Vision-Language Models to Physical Foundation Models}, 
      author={DeepCybo Team and Yu Bin and Haipeng Cao and Zheng Chang and Kai Chen and Youning Chen and Kailin Deng and Yichao Du and Xiaotong Fu and Haoyang Ge and Yunlong Guo and Chenliu Hao and Jiyan He and Xuguo He and Yakun Hou and Kai Hu and Cong Huang and Tuopusen Huang and Yu Huang and Hong Li and Peize Li and Shijie Lian and Xiaopeng Lin and Yun Lin and Haibao Liu and Haochen Liu and Qiuzhi Liu and Shengcai Liu and Zhiqiang Liu and Tao Luo and Peng Ren and Shuo Ren and Chaoyi Ruan and Zhaolong Shen and Yukun Shi and Qiyuan Su and Yuxuan Tian and Yining Wang and Changti Wu and Hao Wu and Xueyin Xu and Ruoqi Yang and Zhaoyang Yang and Hang Yuan and Zhaoyang Zeng and Hanwen Zhang and Ruimeng Zhang and Yao Zhang and Yibo Zhang and Yuxiang Zhang and Zhirui Zhang and Ziyi Zhang and Zubin Zheng and Zishen Zhuang},
      year={2026},
      eprint={2609.14973},
      archivePrefix={arXiv},
      primaryClass={cs.CV},
      url={https://arxiv.org/abs/2609.14973}, 
}
\bibliographystyle{iclr2027_conference}

\clearpage
\appendix

\section{Qualitative Examples of Egocentric Spatial Tracking}
\label{app:qualitative_tracking}
We compare \textsc{Uruqi}$_{\mathrm{Syn}}$-8B, Qwen3.8-27B, and InternVL3-8B on five selected trajectories. The examples illustrate localization during visibility, absence, and reappearance, followed by matched comparisons of \emph{No QA history} and \emph{Model history}. Numbers report horizontal localization errors; dashed circles mark the 0.5\,m threshold. GT scene geometry and poses are used only for visualization. Only selected frames are shown and each prediction uses the full image history available at that step.
\begingroup
\setlength{\floatsep}{5pt}
\setlength{\intextsep}{5pt}
\setlength{\abovecaptionskip}{4pt}
\setlength{\belowcaptionskip}{0pt}
\begin{figure}[!ht]
\centering
\includegraphics[width=\textwidth]{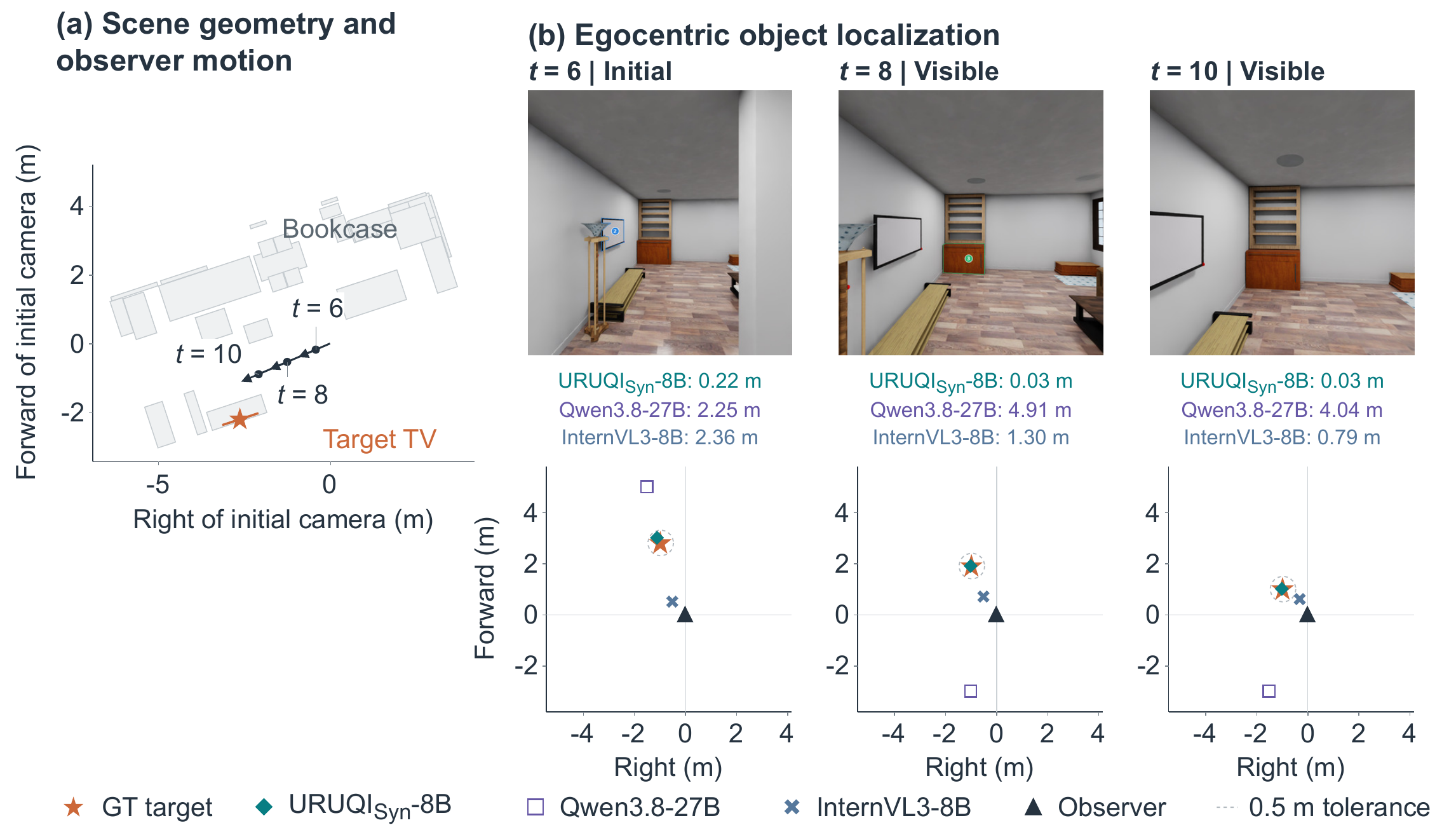}
\caption{\textbf{Localization while the target remains visible.} Under \emph{No QA history}, \textsc{Uruqi}$_{\mathrm{Syn}}$-8B localizes the TV across changing viewpoints, with errors of 0.22, 0.03, and 0.03\,m.}
\label{fig:tracking_tv}
\end{figure}
\begin{figure}[!ht]
\centering
\includegraphics[width=\textwidth]{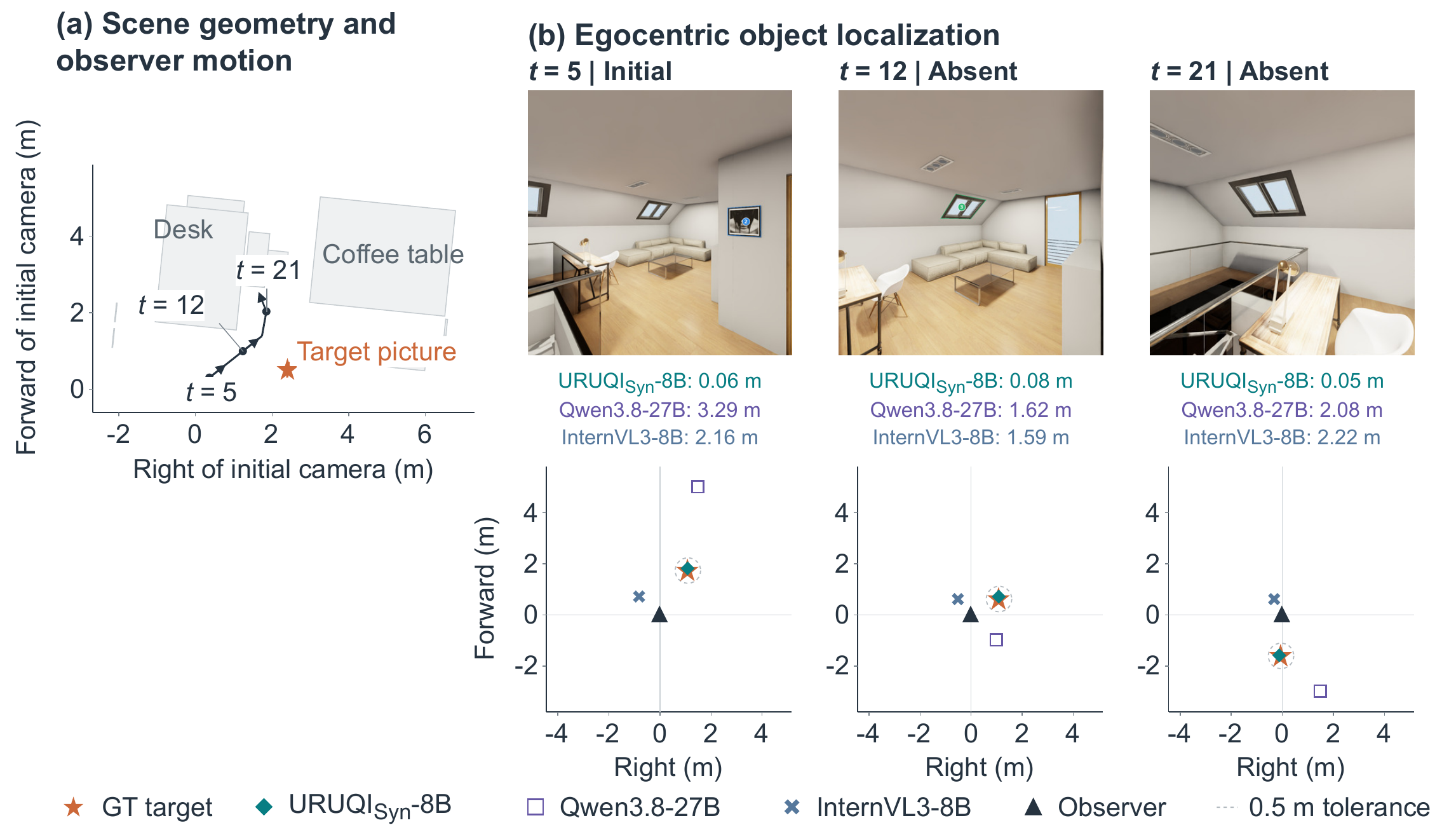}
\caption{\textbf{Localization after the target leaves view.} Under \emph{No QA history}, the picture is absent at the two later observations; \textsc{Uruqi}$_{\mathrm{Syn}}$-8B maintains low errors of 0.08 and 0.05\,m in this selected trajectory.}
\label{fig:tracking_house}
\end{figure}
\clearpage
\begin{figure}[!ht]
\centering
\includegraphics[width=\textwidth]{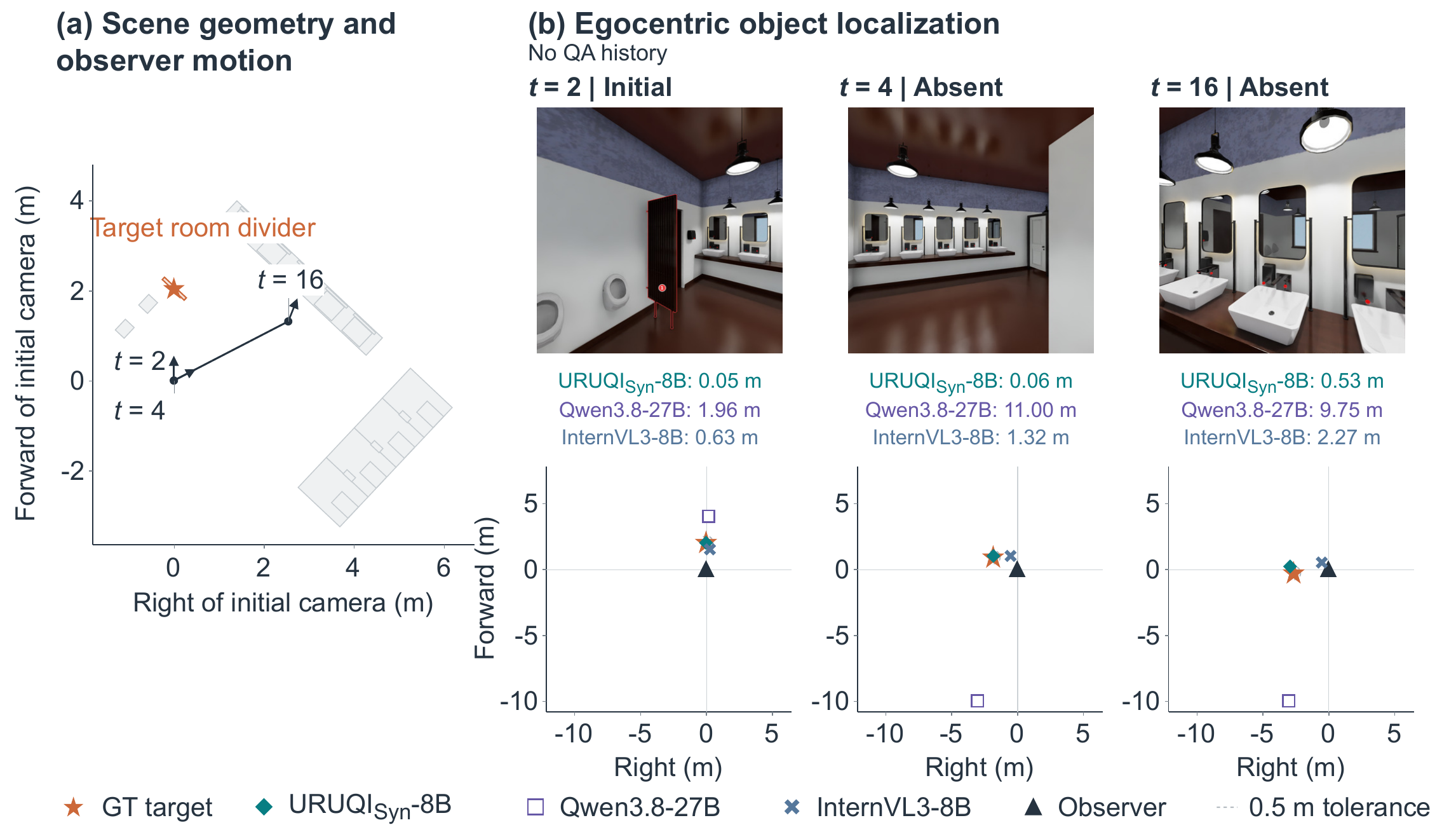}
\par\vspace{12pt}
\includegraphics[width=\textwidth]{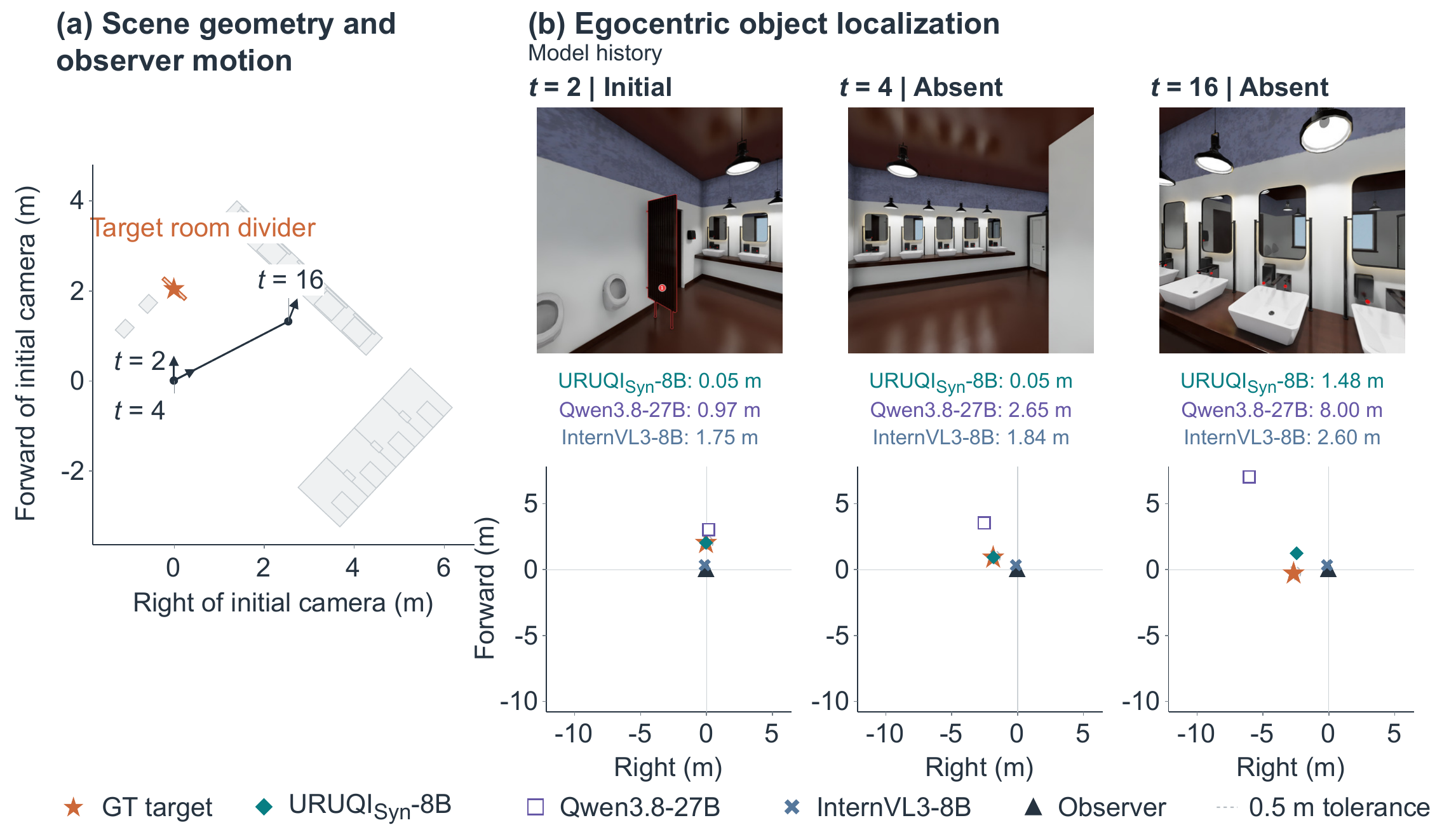}
\caption{\textbf{Localization error during object absence.} The room divider is absent at $t=4$ and $t=16$. At $t=16$, \textsc{Uruqi}$_{\mathrm{Syn}}$-8B has an error of 0.53\,m under No QA history and 1.48\,m under Model history. Top: \emph{No QA history}; bottom: \emph{Model history}. Both conditions use the same object and observation steps, with identical coordinate limits across all six localization plots.}
\label{fig:history_pair_1}
\end{figure}
\clearpage
\begin{figure}[!ht]
\centering
\includegraphics[width=\textwidth]{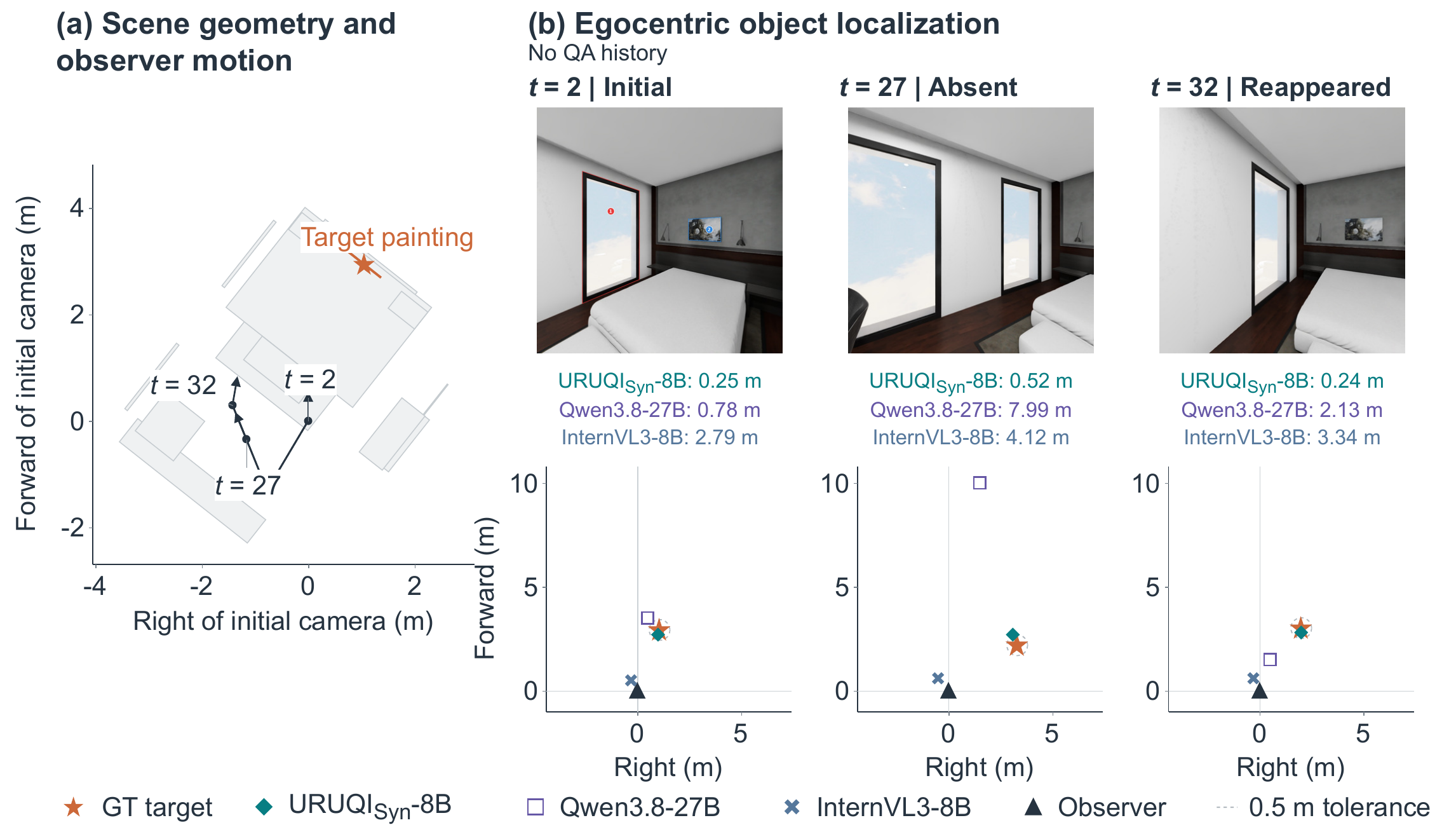}
\par\vspace{12pt}
\includegraphics[width=\textwidth]{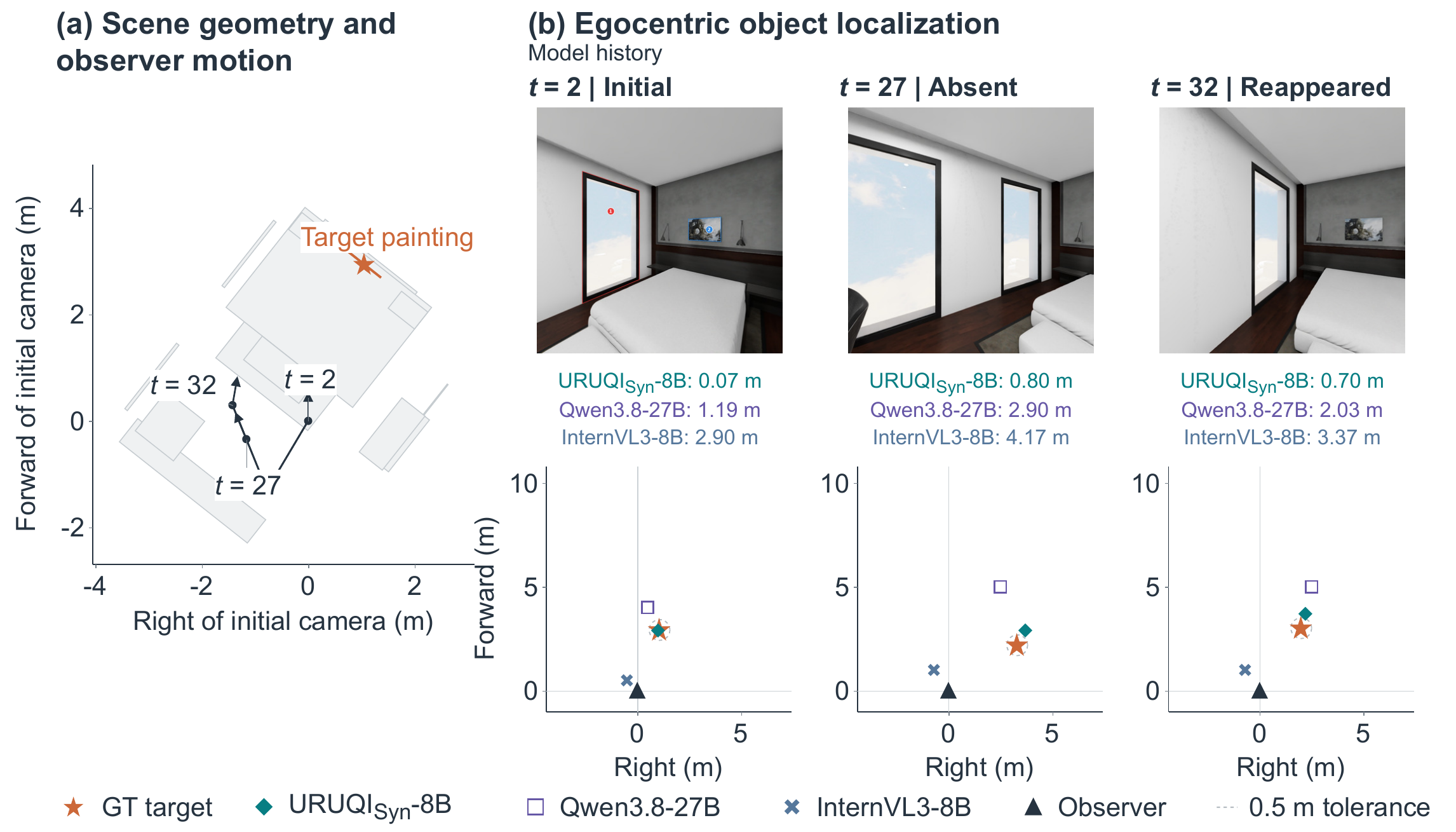}
\caption{\textbf{Localization after object reappearance.} The painting is absent at $t=27$ and reappears at $t=32$. For \textsc{Uruqi}$_{\mathrm{Syn}}$-8B, error decreases from 0.52 to 0.24\,m under No QA history and from 0.80 to 0.70\,m under Model history. Top: \emph{No QA history}; bottom: \emph{Model history}. Both conditions use the same object and observation steps, with identical coordinate limits across all six localization plots.}
\label{fig:history_pair_2}
\end{figure}
\clearpage
\begin{figure}[!ht]
\centering
\includegraphics[width=\textwidth]{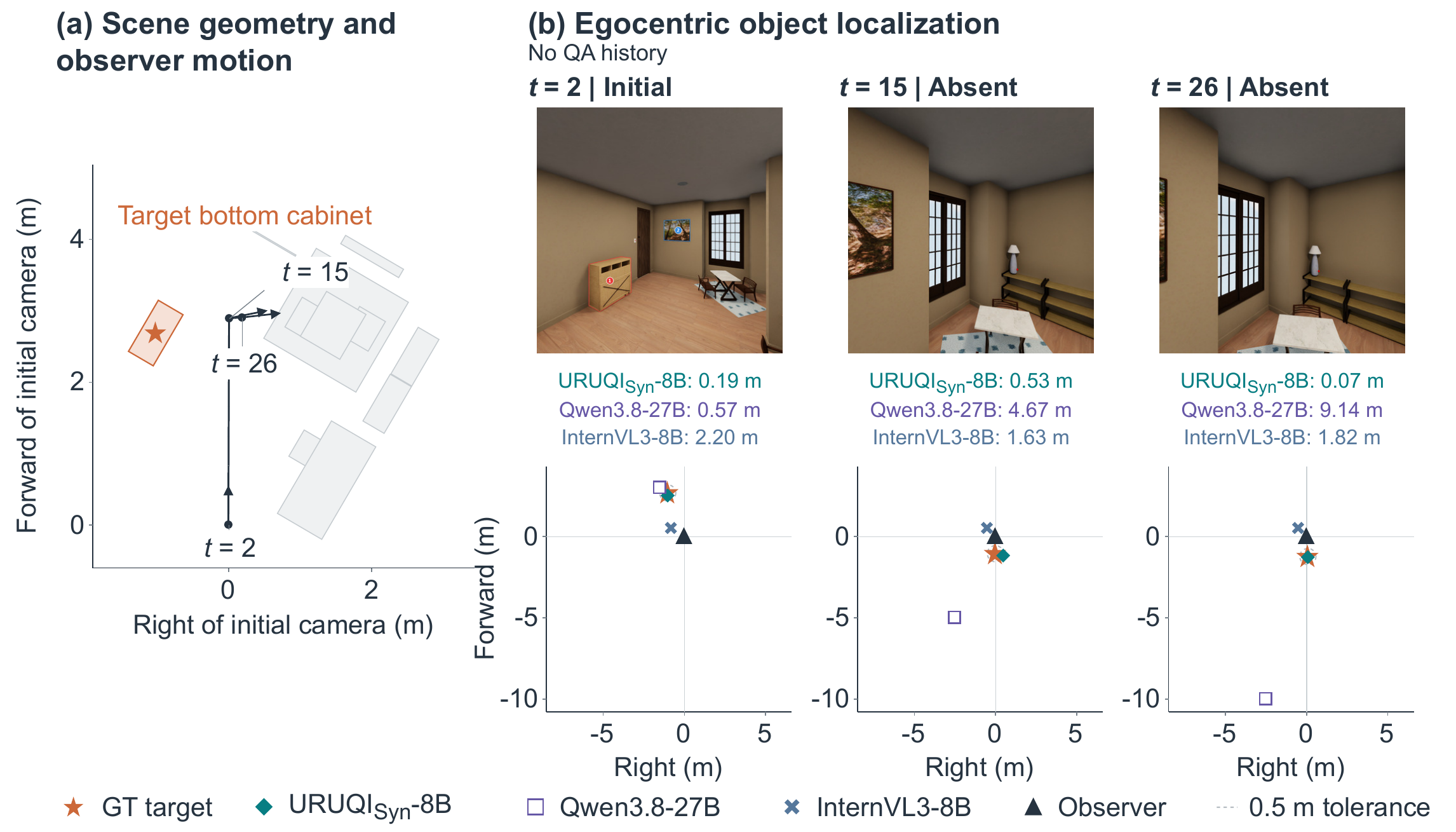}
\par\vspace{12pt}
\includegraphics[width=\textwidth]{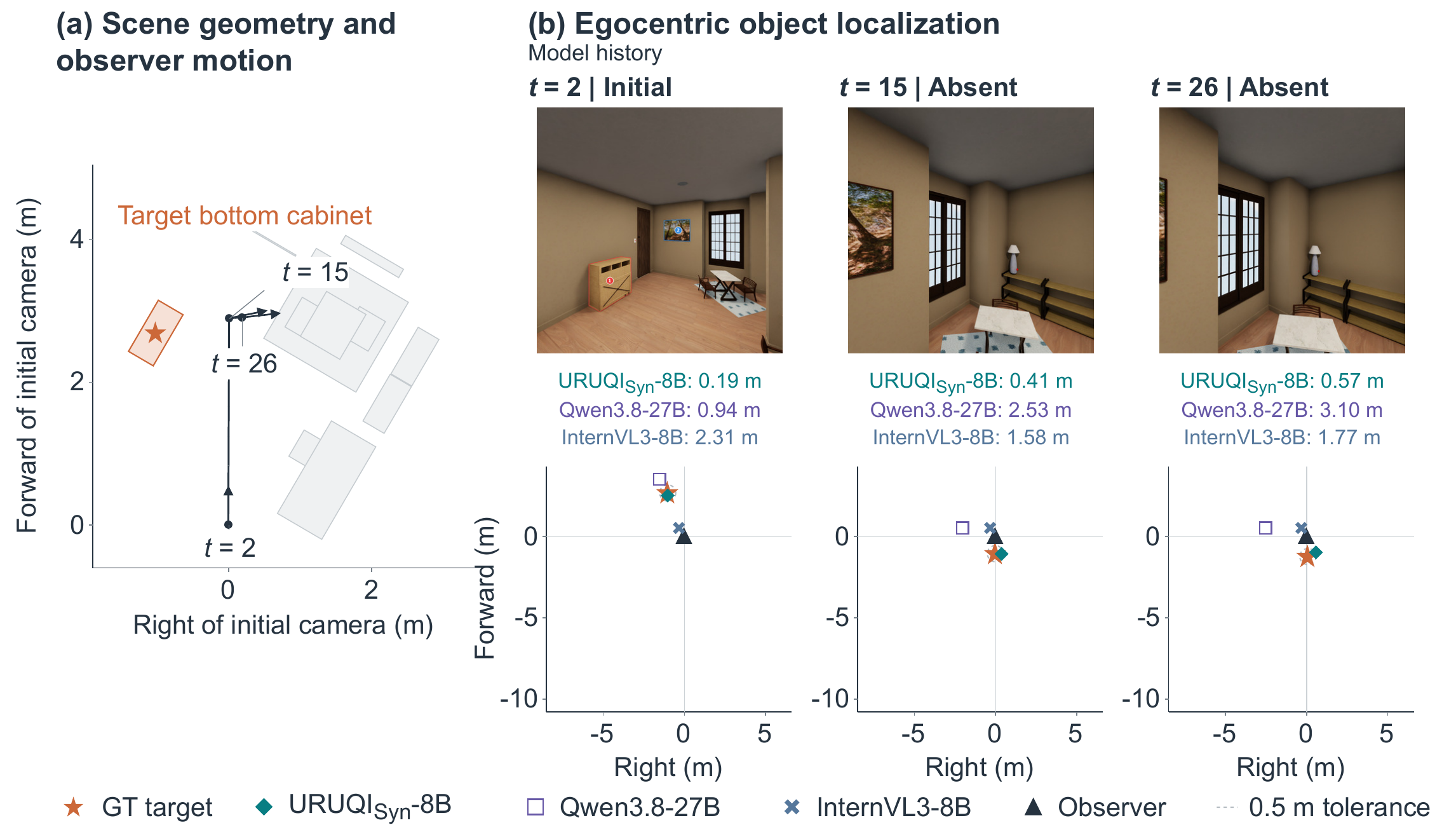}
\caption{\textbf{Variation in the effect of answer history.} For the absent cabinet, \textsc{Uruqi}$_{\mathrm{Syn}}$-8B has lower error under Model history at $t=15$ (0.41 versus 0.53\,m), but higher error at $t=26$ (0.57 versus 0.07\,m). Top: \emph{No QA history}; bottom: \emph{Model history}. Both conditions use the same object and observation steps, with identical coordinate limits across all six localization plots.}
\label{fig:history_pair_3}
\end{figure}

\clearpage
\endgroup

\section{Spatial Motifs and Trajectory Acquisition}
\label{app:spatial_motifs}

\subsection{Motif Instantiation}
\label{app:motif_instantiation}

\paragraph{Motif definitions.}
We instantiate each motif by selecting scene objects and searching for a camera trajectory that satisfies its motion and visibility constraints. \emph{(1) Rotate in place} bounds displacement from the initial camera position while changing heading. \emph{(2) Straight pass} translates past the object with a bound on cumulative rotation. \emph{(3) Walk and turn} combines translation and heading changes after an initial observation of the target. \emph{(4) Multi-turn walk} separates the motion into translational and turning segments, with constraints on both the number of turns and accumulated rotation.
\emph{(5) Occlusion traversal} moves the observer toward or through an occluding region, with constraints that distinguish physical occlusion from disappearance outside the field of view. \emph{(6) Reference survey} binds an origin object, a facing anchor, and a target, then scans their bearings from a fixed station. Each object must be observed, while no single frame provides a clear view of all three. \emph{(7) Landmark chain} connects two endpoint objects through intermediate anchors. Each adjacent pair must have clear co-visible observations, while the endpoints remain non-co-visible throughout the sequence.

\paragraph{Trajectory search.}
To search for these trajectories, we project scene geometry onto the ground plane to construct an occupancy map and retain object centers, extents, categories, and instance identities for role binding. Camera stations are sampled in free space. Unobstructed station pairs are connected directly; other pairs use grid-based A* search. Routes are simplified and interpolated into camera poses with bounded inter-frame translation and rotation. Candidate trajectories are screened for camera clearance, path validity, and the constraints of the selected motif.

\paragraph{Acquisition settings.}
The default trajectory-planning configuration uses a camera height of 1.5\,m and a horizontal field of view of 90\(^{\circ}\) for geometric planning. Consecutive planned observations are constrained to at most 1.2\,m of translation and 40\(^{\circ}\) of rotation. Camera clearance is checked with a 0.30\,m horizontal radius over heights of 0.10--1.70\,m. The planned poses are exported as an ordered camera schedule, with one observation associated with each scheduled viewpoint. Sequence length is therefore measured in observation steps.

\subsection{Trajectory Rendering and Validation}
\label{app:motif_certification}

Candidate trajectories are rendered in OmniGibson~\citep{li2022behavior}, recording RGB, camera pose, metric depth, and instance and semantic identity channels at each observation. Geometric visibility estimates guide trajectory search; rendered instance masks establish object support in the resulting images. The evidence checks evaluate the required visibility transitions and co-visibility relations. For occlusion-traversal candidates, depth and instance evidence are used to check whether disappearance is caused by a foreground blocker. Object bindings, poses, and supporting observations are retained for compilation.

\subsection{Trajectory Statistics}
\label{app:trajectory_statistics}

The source pool for \name{} contains 11,738 distinct trajectories across 50 BEHAVIOR scenes~\citep{li2022behavior}, with 212,209 rendered observations in total. Trajectories contain 6--52 observations, with a mean of 18.08 and a median of 15. Each source trajectory yields multiple QA episodes by varying the selected observations, target objects, and queries. 
\paragraph{Data splits.}
We partition the source trajectories into 8,818 training, 2,696 test, and 224 validation trajectories. The 12 test scenes are disjoint from the 38 training scenes. The final benchmark uses eight of the 12 held-out test scenes.

\section{Spatial Supervision Compilation}
\label{app:supervision_compilation}

\subsection{Coordinate Conventions and Motion Targets}
\label{app:coordinate_conventions}

\paragraph{Reference frames.}
Positions are expressed in meters and reported angles in degrees. The world frame is right-handed with $+Z$ upward. The pose $T_t$ maps observer-frame coordinates to world coordinates; therefore, $T_i^{-1}T_j$ maps coordinates from frame $j$ to frame $i$.

For planar motion targets and horizontal object-location targets, we use forward--left coordinates: forward and left are positive, while backward and right are negative. At zero yaw, the observer faces world $+Y$, with world $+X$ to its right. For yaw $\psi_t$, the forward and left basis vectors form the matrix
\begin{equation}
B_t =
\begin{bmatrix}
-\sin\psi_t & -\cos\psi_t \\
 \cos\psi_t & -\sin\psi_t
\end{bmatrix},
\end{equation}
which maps observer-relative horizontal coordinates to world-plane coordinates.

\paragraph{Self-motion targets.}
Let $o_t\in\mathbb{R}^2$ denote the observer's horizontal position. For observations $i<j$, the displacement and heading targets are
\begin{equation}
d_{ij}=B_i^{\mathsf T}(o_j-o_i),
\qquad
\delta\psi_{ij}=\operatorname{wrap}(\psi_j-\psi_i),
\end{equation}
where $\operatorname{wrap}$ returns the signed principal angle. The two components of $d_{ij}$ give displacement along the forward and left axes of the earlier observation. Positive heading change denotes a left turn. These targets describe the net change between the named viewpoints. Traveled distance and accumulated rotation are computed separately by summing translation magnitudes and absolute heading changes over the intervening observed steps.

\paragraph{Object coordinates under observer motion.}
For a static object with world bounding-box center $c$, let $c_{xy}$ denote its horizontal projection. Its horizontal location at step $t$ is
\begin{equation}
z_t=B_t^{\mathsf T}(c_{xy}-o_t).
\end{equation}
The same location expressed at a later viewpoint satisfies
\begin{equation}
z_j=B_j^{\mathsf T}B_i(z_i-d_{ij}).
\label{eq:app_state_update}
\end{equation}
This relation expresses the same static object in the observer frames before and after motion. Object-location targets are computed directly from scene geometry.

\subsection{Object Identity and Persistent Localization}
\label{app:object_state_targets}

\paragraph{Establishing object identity.}
Persistent object-localization queries refer to objects clearly observed in the available image history. Each question identifies its target by an unambiguous category name or by a marker in an earlier image. Scene instance identities associate these references across viewpoints, while rendered visibility evidence determines whether the object is currently visible or supported only by earlier observations. This ensures that queries about objects outside the current view refer to entities already encountered in the episode.

\paragraph{Localization across viewpoints.}
We construct sequences of location queries for the same object as the observer moves, including periods when the object is no longer visible and, when available, its subsequent reappearance. Each target is computed from the object's fixed world center and the camera pose at the queried viewpoint, following Eq.~\ref{eq:app_state_update}. These sequences supervise how an established object's observer-relative location changes with self-motion, even when no new visual observation of that object is available.

\subsection{Operations over Spatial State}
\label{app:state_operations}

\paragraph{Operation families.}
M3 supervision applies spatial operations to information established along a trajectory. We organize these operations into six broad families, described in Table~\ref{tab:state_operations}. 

\begin{table}[htbp]
\caption{Examples of spatial operations used in M3 supervision.}
\label{tab:state_operations}
\centering
\small
\renewcommand{\arraystretch}{1.18}
\begin{tabularx}{\linewidth}{@{}>{\raggedright\arraybackslash}p{0.25\linewidth}>{\raggedright\arraybackslash}X>{\raggedright\arraybackslash}X@{}}
\toprule
Operation family & Operation & Examples \\
\midrule
Reference-frame reasoning & Express relations in a specified frame & Camera-relative direction; object-centered reference \\
Hypothetical motion & Update relations after imagined motion & Rotation; translation; turn--move sequences \\
Inverse spatial inference & Solve for an unknown from spatial constraints & Target identification from directional constraints \\
Measurement and comparison & Measure, compare, or rank geometry & Distance; size; height; area \\
Temporal and set operations & Query events and combine observations & Appearance order; reappearance; counts; set overlap \\
Route reasoning & Evaluate paths between locations & Route description; traversable-length comparison \\
\bottomrule
\end{tabularx}
\end{table}

\section{Dense Trajectory Evaluation Details}
\label{app:dense_evaluation}

\subsection{Evaluation Protocol}
\label{app:dense_protocol}

\paragraph{Object registration.}
An object is registered after two consecutive observations satisfying the following conditions: its center is 0.5--5\,m from the camera horizontally, its instance mask contains at least 4,000 pixels and lies at least 16 pixels from the image boundary, the mask covers at least 60\% of the projected bounding-box area, and at most 5\% of the projected box lies outside the image. Pixel thresholds are normalized to a $1024\times1024$ image. Mask coverage serves as a visibility proxy.

We register at most three objects per trajectory in chronological order. Registration uses only current and past observations. A numbered marker identifies each object in its registration image, which subsequent questions reference. Once registered, an object remains eligible for queries even when it becomes invisible or leaves the registration distance range.

\paragraph{Queries and visibility stages.}
At each frame after the first, we query motion from the preceding frame, followed by the current observer-relative locations of all registered objects. Location targets are the horizontal camera-frame coordinates of the scene object bounding-box centers. Of the 100 trajectories, 90 contain registered objects and 87 include absent-object queries; the other ten contribute only self-motion queries.

Table~\ref{tab:dense_visibility_stages} defines the five mutually exclusive stages used to group localization queries. Overall localization scores include all five stages.

\begin{table}[htbp]
\caption{Localization query stages in the dense evaluation. Pixel counts are normalized to $1024\times1024$ resolution; stages other than Initial apply after registration.}
\label{tab:dense_visibility_stages}
\centering
\small
\renewcommand{\arraystretch}{1.15}
\begin{tabularx}{\linewidth}{@{}l>{\raggedright\arraybackslash}Xr@{}}
\toprule
Stage & Definition & Queries \\
\midrule
Initial & Query at the object's registration frame & 231 \\
Absent & Zero visible instance pixels & 2,240 \\
Reappeared & First observation with at least 256 instance pixels after an absence & 79 \\
Visible & Other observations with at least 256 instance pixels after registration & 1,651 \\
Weak visibility & Positive instance-pixel count below 256 & 6 \\
\midrule
Total & & 4,207 \\
\bottomrule
\end{tabularx}
\end{table}

\subsection{Baseline Prompts}
\label{app:baseline_prompts}

For InternVL3-8B and Qwen3.8-27B, we append answer-format instructions to each motion and object-localization question.

\paragraph{Self-motion estimation.}
\begin{quote}
\small
\textbf{User.} For this question, compare Image 1 with the current image.

How do your position and heading change from Image 1 to the current image?

Answer briefly with estimated numeric distances in meters and the heading change in degrees, relative to the earlier camera. Use forward/backward, left/right, and turn left/right. Format example only: Move 0.4 m forward and 0.2 m right. Turn left 15 degrees. For no change, answer: Stay still. Give your best estimate; output only the short answer, no explanation.

\textbf{InternVL3-8B:} Move 0.3 m forward. Turn right 10 degrees.

\textbf{Qwen3.8-27B:} Stay still.

\end{quote}

\paragraph{Object localization.}
\begin{quote}
\small
\textbf{User.} Using observations through Image 2, where is the object introduced as Marker 1 in Image 2 relative to your position and heading in Image 2? Give forward/backward and left/right distances in meters.

Answer briefly with two estimated distances in meters in the current camera frame, using forward/backward and left/right. Format example only: 2.0 m forward, 0.5 m left. Give your best estimate; output only the short answer, no explanation.

\textbf{InternVL3-8B:} 1.5 m forward, 0.5 m right.

\textbf{Qwen3.8-27B:} 3.0 m forward, 1.5 m right

\end{quote}

\subsection{Metrics and Geometric Propagation}
\label{app:dense_metrics}

\paragraph{Scoring.}
Object locations use the horizontal observer frame defined in Appendix~\ref{app:coordinate_conventions}. For a query set $Q$, let $V\subseteq Q$ contain the queries with valid location estimates. Position error and correctness are
\begin{equation}
e_i=\|\hat{z}_i-z_i\|_2,
\qquad
c_i=\begin{cases}
\mathbf{1}[e_i\leq0.5\,\mathrm{m}], & i\in V,\\
0, & i\notin V.
\end{cases}
\end{equation}
Missing or unparseable predictions, including uninitialized propagated states, count as incorrect. Accuracy and mean position error are
\begin{equation}
\mathrm{Acc}_{\mathrm{query}}=\frac{100}{|Q|}\sum_{i\in Q}c_i,
\qquad
\overline{e}=\frac{1}{|V|}\sum_{i\in V}e_i.
\end{equation}
All localization results weight queries equally. Accuracy includes every query, while mean error is computed over valid estimates. The geometric conditions in Table~\ref{tab:pose-assisted-full} provide estimates for all queries.

For self-motion, translation error is the Euclidean difference between predicted and ground-truth horizontal displacements in the earlier camera frame. For motion query $i$, the predicted and ground-truth heading changes are $\widehat{\delta\psi}_i$ and $\delta\psi_i$; their error is $|\operatorname{wrap}_{[-180^\circ,180^\circ)}(\widehat{\delta\psi}_i-\delta\psi_i)|$. Motion errors are averaged over valid outputs.

\paragraph{Geometric propagation.}
An object is initialized at the first step $t_0$ with a positive model visibility response and a valid location estimate $\hat{z}_{t_0}$. Using the horizontal basis $B_t$ and observer position $o_t$ from Appendix~\ref{app:coordinate_conventions}, we store a fixed world-plane point and compute its subsequent observer-relative locations as
\begin{equation}
\hat{c}_{xy}=B_{t_0}\hat{z}_{t_0}+o_{t_0},
\qquad
\hat{z}^{\mathrm{prop}}_t=B_t^{\mathsf T}(\hat{c}_{xy}-o_t).
\end{equation}
Later VLM location estimates do not update the stored point. Initialization uses model predictions without ground-truth instance masks; direct prediction requires no visibility gate. The pose-source comparison holds object-location and auxiliary visibility predictions fixed and uses the same propagation rule, varying only whether poses are ground truth or integrated from predicted motion. Ground-truth poses are supplied only to the geometry module.

\subsection{Complete Results}
The following table supplements Table~\ref{tab:pose-assisted-unified} with overall accuracy and mean position errors under the same evaluation conditions.

\begin{table*}[t]
\centering
\caption{\textbf{Complete pose-assisted object-localization results.}
(a) Direct prediction and pose-based propagation use the same object predictions
under No QA history; GT poses are used for propagation.
(b) With Model-history object predictions and pose-based propagation fixed,
we vary the source of camera poses.
GT poses are supplied only to the geometry module, not to the VLM.
Acc@0.5m uses a 0.5\,m horizontal-error threshold in the current camera frame;
all metrics weight queries equally.
All and Absent contain 4,207 and 2,240 queries, respectively.
Bold indicates the best result within each model and panel.}
\label{tab:pose-assisted-full}
\begingroup
\small
\setlength{\tabcolsep}{5pt}
\renewcommand{\arraystretch}{1.12}

\textbf{(a) Direct prediction versus pose-based propagation}\\[2pt]
\emph{Fixed: No QA history; GT poses for propagation.}\\[4pt]
\begin{tabular*}{\linewidth}{@{\extracolsep{\fill}}llrrr@{}}
\toprule
\multirow{2}{*}{Model}
& \multirow{2}{*}{Localization method}
& \multicolumn{2}{c}{Acc@0.5m (\%) $\uparrow$}
& \multirow{2}{*}{\shortstack{Mean error (m) $\downarrow$\\All}} \\
\cmidrule(lr){3-4}
& & All & Absent & \\
\midrule
\multirow{2}{*}{Qwen3.8-27B}
& Direct prediction & 6.44 & 1.29 & 4.725 \\
& Pose-based propagation & \textbf{16.81} & \textbf{16.88} & \textbf{1.322} \\
\midrule
\multirow{2}{*}{\textsc{URUQI}$_{\mathrm{Syn}}$-8B}
& Direct prediction & 49.56 & 35.13 & 0.743 \\
& Pose-based propagation & \textbf{69.65} & \textbf{71.61} & \textbf{0.412} \\
\bottomrule
\end{tabular*}

\vspace{9pt}
\textbf{(b) Effect of camera-pose source}\\[2pt]
\emph{Fixed: Model history; pose-based propagation. Evaluated on absent objects.}\\[4pt]
\begin{tabular*}{\linewidth}{@{\extracolsep{\fill}}llrr@{}}
\toprule
Model & Camera-pose source
& \shortstack{Acc@0.5m (\%) $\uparrow$\\Absent}
& \shortstack{Mean error (m) $\downarrow$\\Absent} \\
\midrule
\multirow{2}{*}{Qwen3.8-27B}
& GT poses & \textbf{18.48} & \textbf{0.961} \\
& Integrated predicted motion & 3.97 & 3.737 \\
\midrule
\multirow{2}{*}{\textsc{URUQI}$_{\mathrm{Syn}}$-8B}
& GT poses & \textbf{72.86} & \textbf{0.397} \\
& Integrated predicted motion & 34.02 & 1.084 \\
\bottomrule
\end{tabular*}
\endgroup
\end{table*}

\subsection{Auxiliary Visibility and Object Correspondence}
\label{app:visibility_check}

\paragraph{Diagnostic protocol.}
For each localization query, we separately ask whether the registered object is currently visible and, if so, request a pixel on that object. The model receives all images up to the current frame, including the marked registration image, without QA history or privileged geometry. The same predictions are reused across the model's history and pose conditions. A valid positive response includes an in-bounds pixel coordinate; invalid or uncertain responses cannot initialize propagation. Instance masks are used only to score pixel correspondence, not to accept an initialization. This auxiliary query does not affect the native localization scores in Table~\ref{tab:atomic-spatial}.

\paragraph{Metrics.}
Ground-truth visibility requires at least one target-instance pixel, giving 1,967 visible and 2,240 absent queries. Precision and recall measure valid positive visibility responses against these labels. False-visible rate is the fraction of absent queries receiving a valid positive response. Pixel Hit is the fraction of all ground-truth-visible queries with a valid positive response and a predicted pixel on the correct instance; missed detections and invalid responses receive no credit.

\begin{table}[htbp]
\centering
\small
\setlength{\tabcolsep}{5pt}
\caption{Auxiliary visibility and object correspondence on 4,207 queries. All values are percentages; bold marks the best value in each column.}
\label{tab:visibility-correspondence}
\begin{tabular}{@{}lrrrr@{}}
\toprule
Model & \shortstack{Visibility\\precision $\uparrow$}
& \shortstack{Visibility\\recall $\uparrow$}
& \shortstack{Pixel Hit\\$\uparrow$}
& \shortstack{False-visible\\rate $\downarrow$} \\
\midrule
InternVL3-8B & 54.13 & 88.87 & 7.52 & 66.12 \\
Qwen3.8-27B & \textbf{93.51} & 78.39 & 50.64 & \textbf{4.78} \\
\textsc{URUQI}$_{\mathrm{Syn}}$-8B & 68.31 & \textbf{99.49} & \textbf{71.68} & 40.54 \\
\bottomrule
\end{tabular}
\end{table}

\paragraph{Results.}
Table~\ref{tab:visibility-correspondence} shows that \textsc{URUQI}$_{\mathrm{Syn}}$-8B achieves the highest Pixel Hit, but has a higher false-visible rate than Qwen3.8-27B (40.54\% versus 4.78\%). These metrics are computed over their respective query sets, not only at initialization. As described in Appendix~\ref{app:dense_metrics}, visibility responses after initialization do not change the stored object location.

\clearpage

\end{document}